\pdfoutput=1

\documentclass[11pt]{article}

\usepackage{acl}
\usepackage[parfill]{parskip}
\usepackage[switch]{lineno} 
\usepackage{amsmath}

\usepackage{times}
\usepackage{latexsym}
\usepackage{float}
\usepackage[T1]{fontenc}

\usepackage[utf8]{inputenc}
\usepackage{graphicx}

\usepackage{microtype}
\usepackage{subfigure}
\usepackage{booktabs} 

\usepackage{hyperref}
\usepackage{balance}

\usepackage{amssymb}
\usepackage{mathtools}
\usepackage{amsthm}

\usepackage[normalem]{ulem}
\usepackage{bbm}
\usepackage{bbding}
\usepackage[skip=-1pt]{caption}
\usepackage{makecell}
\usepackage{amsfonts}
\usepackage[inline]{enumitem}
\usepackage{multirow}
\usepackage{multicol}
\usepackage{stmaryrd}
\usepackage{longtable}
\usepackage{tabularx} 
\usepackage{pifont}   
\usepackage{tikz}
\usepackage{tcolorbox}
\usepackage{mdframed}

\usepackage{lipsum}
\usepackage{url}

\usepackage[capitalize,noabbrev]{cleveref}

\theoremstyle{plain}

\theoremstyle{definition}

\theoremstyle{remark}

\usepackage[textsize=tiny]{todonotes}

\graphicspath{ {./fig/}  }

\newcommand{\new}[1]{\textcolor{black}{#1}}

\DeclareMathOperator*{\argmax}{arg\,max} 

\makeatletter
\DeclareRobustCommand\onedot{\futurelet\@let@token\@onedot}
\def\@onedot{\ifx\@let@token.\else.\null\fi\xspace}

\makeatother

\usepackage{listings}
\usepackage{xcolor}

\colorlet{punct}{red!60!black}
\definecolor{background}{HTML}{EEEEEE}
\definecolor{delim}{RGB}{20,105,176}
\colorlet{numb}{magenta!60!black}

\lstdefinelanguage{json}{
    basicstyle=\normalfont\ttfamily,
    showstringspaces=false,
    breaklines=true,
    frame=lines,
    backgroundcolor=\color{background},
    literate=
     *{0}{{{\color{numb}0}}}{1}
      {1}{{{\color{numb}1}}}{1}
      {2}{{{\color{numb}2}}}{1}
      {3}{{{\color{numb}3}}}{1}
      {4}{{{\color{numb}4}}}{1}
      {5}{{{\color{numb}5}}}{1}
      {6}{{{\color{numb}6}}}{1}
      {7}{{{\color{numb}7}}}{1}
      {8}{{{\color{numb}8}}}{1}
      {9}{{{\color{numb}9}}}{1}
      {:}{{{\color{punct}{:}}}}{1}
      {,}{{{\color{punct}{,}}}}{1}
      {\{}{{{\color{delim}{\{}}}}{1}
      {\}}{{{\color{delim}{\}}}}}{1}
      {[}{{{\color{delim}{[}}}}{1}
      {]}{{{\color{delim}{]}}}}{1},
}    

\title{Learning to Extract Structured Entities Using Language Models}

\author{
  Haolun Wu\textsuperscript{1, 2}\thanks{\quad Equal contribution with random order.}, 
  Ye Yuan\textsuperscript{1, 2}\footnotemark[1],
  Liana Mikaelyan\textsuperscript{3},
  Alexander Meulemans\textsuperscript{4},\\
  \textbf{Xue Liu\textsuperscript{1, 2}},
  \textbf{James Hensman\textsuperscript{3}},
  \textbf{Bhaskar Mitra\textsuperscript{3}}
  \vspace{1mm}
  \\
  \textsuperscript{1}~McGill University,
  \textsuperscript{2}~Mila - Quebec AI Institute,
  \textsuperscript{3}~Microsoft Research,
  \textsuperscript{4}~ETH Zürich.
  \vspace{1mm}
  \\
  \texttt{\{haolun.wu, ye.yuan3\}@mail.mcgill.ca},\\
  \texttt{xueliu@cs.mcgill.ca},
  \texttt{ameulema@ethz.ch},\\
  \texttt{\{lmikaelyan, jameshensman, bhaskar.mitra\}@microsoft.com}.
}

\begin{document}

\maketitle

\begin{abstract}
Recent advances in machine learning have significantly impacted the field of information extraction, with Language Models (LMs) playing a pivotal role in extracting structured information from unstructured text. Prior works typically represent information extraction as triplet-centric and use classical metrics such as precision and recall for evaluation. We reformulate the task to be entity-centric, enabling the use of diverse metrics that can provide more insights from various perspectives. We contribute to the field by introducing Structured Entity Extraction and proposing the Approximate Entity Set OverlaP (AESOP) metric, designed to appropriately assess model performance. Later, we introduce a new Multistage Structured Entity Extraction (MuSEE) model that harnesses the power of LMs for enhanced effectiveness and efficiency by decomposing the extraction task into multiple stages. Quantitative and human side-by-side evaluations confirm that our model outperforms baselines, offering promising directions for future advancements in structured entity extraction. 
Our source code and datasets are available at \href{https://github.com/microsoft/Structured-Entity-Extraction}{https://github.com/microsoft/Structured-Entity-Extraction}.

\end{abstract}
\section{Introduction}
\label{sec:intro}
\vspace{-2mm}
Information extraction refers to a broad family of challenging natural language processing (NLP) tasks that aim to extract structured information from unstructured text~\citep{cardie1997empirical, eikvil1999information, chang2006survey, sarawagi2008information, grishman2015information, niklaus2018survey, nasar2018information, wang2018clinical, martinez2020information}.
Examples of information extraction tasks include:
\begin{enumerate*}[label=\textit{(\roman*)}]
    \item Named-entity recognition~\citep{ li2020survey},
    \item relation extraction~\citep{kumar2017survey},
    \item event extraction~\citep{li2022survey}, and
    \item coreference resolution~\citep{ stylianou2021neural, liu2023brief},
\end{enumerate*}
as well as higher-order challenges, such as automated knowledge base (KB) and knowledge graph (KG) construction from text~\citep{weikum2010information, ye2022generative, zhong2023comprehensive}.
The latter may in turn necessitate solving a combination of the former more fundamental extraction tasks as well as require other capabilities like entity linking~\citep{shen2014entity, shen2021entity, oliveira2021towards, sevgili2022neural}.

\begin{figure}[t]
    \centering
    \includegraphics[width=1.0\linewidth]{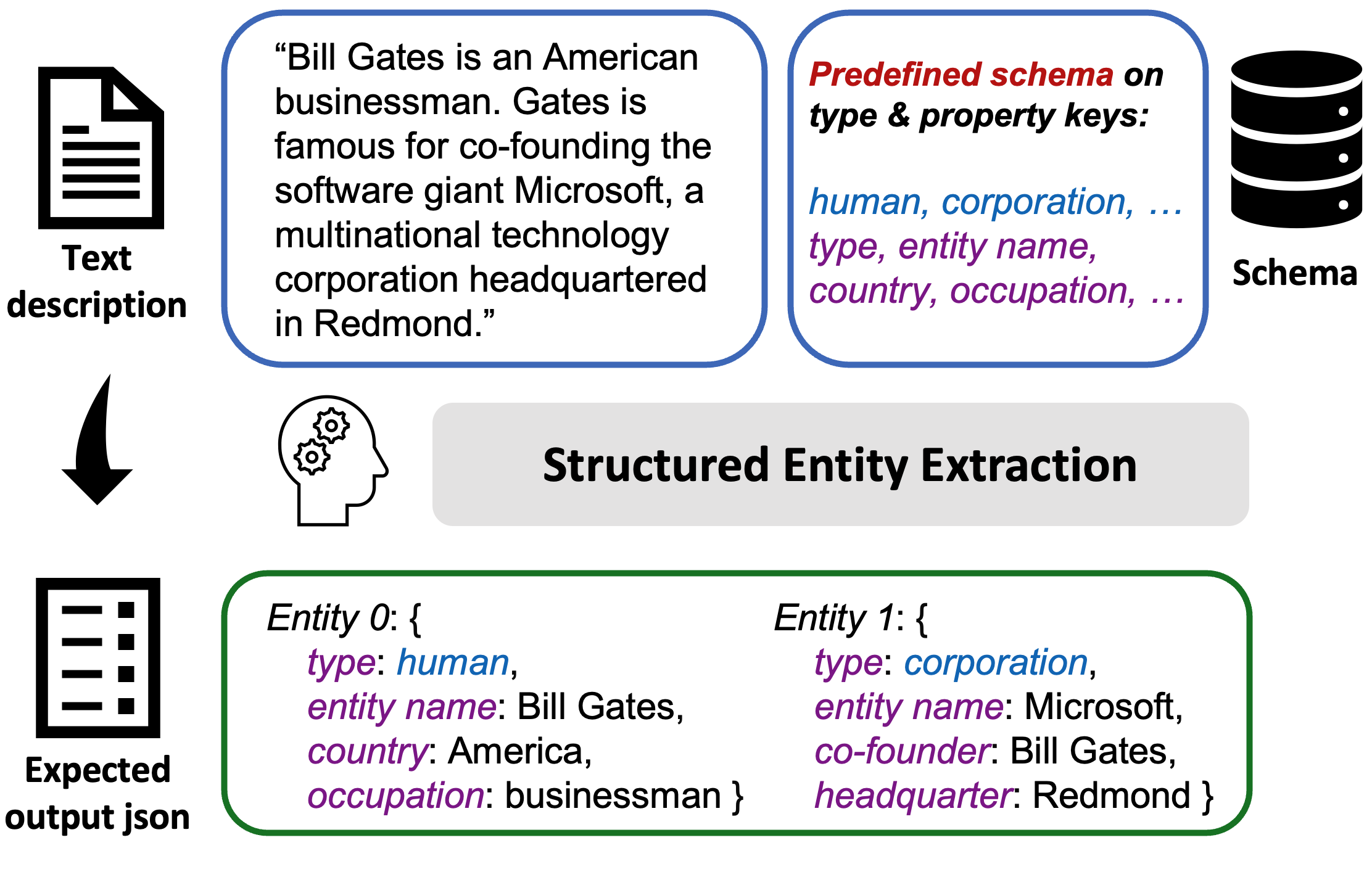}
    \caption{Illustration of the structured entity extraction, an entity-centric formulation of information extraction. Given a text description as well as some predefined schema containing all the candidates of entity types and property keys, we aim to output a structured json for all entities in the text with their information.}
    \label{fig:task}
    \vspace{-6mm}
\end{figure}

Previous formulations and evaluations of information extraction have predominantly centered around the extraction of \textit{$\langle$subject, relation, object$\rangle$} triplets. 
The conventional metrics used to evaluate triplet-level extraction, such as recall and precision, however, might be insufficient to represent a model's understanding of the text from a holistic perspective.
For example, consider a paragraph that mentions ten entities, where one entity is associated with 10 relations as the subject, while each of the other nine entities is associated with only 1 relation as the subject.
Imagine a system that accurately predicts all ten triplets for the heavily linked entity but overlooks the other entities. 
Technically, this system achieves a recall of more than 50\% (i.e., 10 out of 19) and a precision of 100\%. 
However, when compared to another system that recognizes one correct triplet for each of the ten entities and achieves the same recall and precision, it becomes evident that both systems, despite showing identical evaluation scores, offer significantly different insights into the text comprehension.
Moreover, implementing entity-level normalization within traditional metrics is not always easy due to challenges like coreference resolution~\citep{ stylianou2021neural, liu2023brief}, particularly in scenarios where multiple entities share the same name or lack primary identifiers such as names. 
Therefore, we advocate for alternatives that can offer insights from diverse perspectives.

In this work, we propose \textit{S}tructured \textit{E}ntity \textit{E}xtraction, an entity-centric formulation of (strict) information extraction, which facilitates diverse evaluations.
We define a structured entity as a named entity with associated properties and relationships with other named-entities.
Fig.~\ref{fig:task} shows an illustration of the structured entity extraction.
Given a text description, we aim to first identify the two entities ``\textit{Bill Gates}'' and ``\textit{Microsoft}''.
Then, given some predefined schema on all possible entity types and property keys (referred to as a \textit{strict} setting in our scenario), the exact types, property keys, property values on all identified entities in the text are expected to be predicted, as well as the relations between these two entities (i.e., \textit{Bill Gates} co-founded \textit{Microsoft}).
Such extracted structured entities may be further linked and merged to automatically construct KBs from text corpora.
Along with this, we propose a new evaluation metric, \textit{A}pproximate \textit{E}ntity \textit{S}et \textit{O}verla\textit{P} (AESOP), with numerous variants for measuring the similarity between the predicted set of entities and the ground truth set, which is more flexible to include different level of normalization (see default AESOP in Sec.~\ref{sec:prelim} and other variants in Appendix~\ref{appendix:diff_variants}).

In recent years, deep learning has garnered significant interest in the realm of information extraction tasks. 
Techniques based on deep learning for entity extraction have consistently outperformed traditional methods that rely on features and kernel functions, showcasing superior capability in feature extraction and overall accuracy~\citep{yang2022survey}. 
Building upon these developments, our study employs language models (LMs) to solve structured entity extraction. 
We introduce a \textit{Mu}lti-stage \textit{S}tructured \textit{E}ntity \textit{E}xtraction (MuSEE) model, a novel architecture that enhances both effectiveness and efficiency.
Our model decomposes the entire information extraction task into multiple stages, enabling parallel predictions within each stage for enhanced focus and accuracy.
Additionally, we reduce the number of tokens needed for generation, which further improves the efficiency for both training and inference.
Human side-by-side evaluations show similar results as our AESOP metric, which not only further confirm our model's effectiveness but also validate the AESOP metric.


In summary, our main contributions are:
\vspace{-2mm}
\begin{itemize}
\item We introduce an entity-centric formulation of the information extraction task within a strict setting, where the schema for all possible entity types and property keys is predefined.
\vspace{-2mm}
\item We propose an evaluation metric, \textit{A}pproximate \textit{E}ntity \textit{S}et \textit{O}verla\textit{P} (AESOP), with more flexibility tailored for assessing structured entity extraction.
\vspace{-2mm}
\item We propose a new model leveraging the capabilities of LMs, improving the effectiveness and efficiency for structured entity extraction.
\end{itemize}

\section{Related work}
\label{sec:related}
\vspace{-2mm}
%
%
In this section, we first review the formulation of existing information extraction tasks and the metrics used, followed by a discussion of current methods for solving information extraction tasks.
Information extraction tasks are generally divided into open and closed settings.
Open information extraction (OIE), first proposed by \citet{oie2007banko}, is designed to derive relation triplets from unstructured text by directly utilizing entities and relationships from the sentences themselves, without adherence to a fixed schema.
Conversely, closed information extraction (CIE) focuses on extracting factual data from text that fits into a predetermined set of relations or entities, as detailed by \citet{josifoski2022genie}.
While open and closed information extraction vary, both seek to convert unstructured text into structured knowledge, which is typically represented as triplets. These triplets are useful for outlining relationships but offer limited insight at the entity level.
It is often assumed that two triplets refer to the same entity if their subjects match.
However, this assumption is not always held.
Additionally, the evaluation of these tasks relies on precision, recall, and F$1$ scores at the triplet level.
As previously mentioned, evaluating solely on triplet metrics can yield misleading insights regarding the entity understanding.
Thus, it is essential to introduce a metric that assesses understanding at the entity level through entity-level normalization.
In this work, we introduce the AESOP metric, which is elaborated on in Sec.~\ref{sec:aesop}.

Various strategies have been employed in existing research to address the challenges of information extraction.
TextRunner~\cite{yates-etal-2007-textrunner} initially spearheaded the development of unsupervised methods.
Recent progress has been made with the use of manual annotations and Transformer-based models~\cite{vasilkovsky2022detie, kolluru2020openie6}.
Sequence generation approaches, like IMoJIE~\cite{kolluru2020imojie} and GEN2OIE~\cite{kolluru-etal-2022-alignment}, have refined open information extraction by converting it into a sequence-to-sequence task~\cite{cui2018neural}.
GenIE~\cite{josifoski2022genie} focuses on integrating named-entity recognition, relation extraction, and entity linking within a closed setting where a knowledge base is provided.
Recent work, PIVOINE~\cite{lu2023pivoine}, focuses on improving the language model’s generality to various (or unseen) instructions for open information extraction, whereas our focus is on designing a new model architecture for improving the effectiveness and efficiency of language model’s information extraction in a strict setting. 
\section{Structured Entity Extraction}
\label{sec:prelim}
\vspace{-2mm}
In this section, we first describe the structured entity extraction formulation, followed by detailing the Approximate Entity Set OverlaP (AESOP) metric for evaluation.
We would like to emphasize that structured entity extraction is not an entirely new task, but rather a novel entity-centric formulation of information extraction.

\vspace{-2mm}
\subsection{Task Formulation}
\vspace{-2mm}
Given a document $d$, the goal of structured entity extraction is to generate a set of structured entities $\mathcal{E} = \{e_1, e_2, \ldots, e_n \}$ that are mentioned in the document text.
Each structured entity $e$ is a dictionary of property keys $p \in \mathcal{P}$ and property values $v \in \mathcal{V}$, and let $v_{e,p}$ be the value of property $p$ of entity $e$.
In this work we consider only text properties and hence $\mathcal{V}$ is the set of all possible text property values.
If a property of an entity is common knowledge but does not appear in the input document, it will not be considered in the structured entity extraction. Depending on the particular situation, the property values could be other entities, although this is not always the case.

So, the goal then becomes to learn a function $f: d \to \mathcal{E}'=\{e'_1, e'_2, \ldots, e'_m \}$, and we expect the predicted set $\mathcal{E}'$ to be as close as possible to the target set $\mathcal{E}$, where the closeness is measured by some similarity metric $\Psi(\mathcal{E}', \mathcal{E})$.
Note that the predicted set of entities $\mathcal{E}'$ and the ground-truth set $\mathcal{E}$ may differ in their cardinality, and our definition of $\Psi$ should allow for the case when $|\mathcal{E}'| \neq |\mathcal{E}|$.
Finally, both $\mathcal{E}'$ and $\mathcal{E}$ are unordered sets and hence we also want to define $\Psi$ to be order-invariant over $\mathcal{E}'$ and $\mathcal{E}$.
As we do not need to constrain $f$ to produce the entities in any strict order, it is reasonable for $\Psi$ to assume the most optimistic assignment of $\mathcal{E}'$ with respect to
$\mathcal{E}$.
We denote $\vec{E}'$ and $\vec{E}$ as some arbitrary but fixed ordering over items in prediction set $\mathcal{E}'$ and ground-truth set $\mathcal{E}$ for allowing indexing.
\vspace{-2mm}
\subsection{Approximate Entity Set OverlaP (AESOP) Metric} \label{sec:aesop}
\vspace{-2mm}
We propose a formal definition of the Approximate Entity Set OverlaP (AESOP) metric, which focuses on the entity-level and more flexible to include different level of normalization:

\vspace{-5mm}
\begin{small} \label{eq:aesop}
\begin{align}
    \Psi(\mathcal{E}', \mathcal{E}) = \frac{1}{\mu} \bigoplus_{i,j}^{m,n} {\mathbf{F}_{i,j} \cdot \psi_\text{ent}(\vec{E'}_{{i}}, \vec{E}_j)}, \label{eqn:big-delta}
\end{align}
\end{small} %
which is composed of two phases: \textit{(\romannumeral1)} \textit{optimal entity assignment} for obtaining the assignment matrix $\mathbf{F}$ to let us know which entity in $\mathcal{E}'$ is matched with which one in $\mathcal{E}$, and \textit{(\romannumeral2)} \textit{pairwise entity comparison} through $\psi_\text{ent}(\vec{E'}_i, \vec{E}_j)$, which is a similarity measure defined between any two arbitrary entities $e'$ and $e$.
We demonstrate the details of these two phases in this section.
We implement $\Psi$ as a linear sum $\bigoplus$ over individual pairwise entity comparisons $\psi_\text{ent}$, and $\mu$ is the maximum of the sizes of the target set and the predicted set, i.e., $\mu = \max\{m, n\}$.

\vspace{-2mm}
\paragraph{Phase 1: Optimal Entity Assignment.}
The optimal entity assignment is directly derived from a matrix $\mathbf{F}\in\mathbb{R}^{m\times n}$, which is obtained by solving an assignment problem between $\mathcal{E}'$ and $\mathcal{E}$.
Here, the matrix $\mathbf{F}$ is a binary matrix where each element $\mathbf{F}_{i,j}$ is 1 if the entity $\vec{E}'_i$ is matched with the entity $\vec{E}_j$, and 0 otherwise.
Before formulating the assignment problem, we first define a similarity matrix $\mathbf{S}\in\mathbb{R}^{m\times n}$ where each element $\mathbf{S}_{i,j}$ quantifies the similarity between the $i$-th entity in $\vec{E}'$ and the $j$-th entity in $\vec{E}$ for the assignment phase. For practical implementation, we ensure inclusion of the union set of property keys from both the $i$-th entity in $\vec{E}'$ and the $j$-th entity in $\vec{E}$ for each of these entities. When a property key is absent, its corresponding property value is set to be an empty string. The similarity is then computed as a weighted average of the Jaccard index~\citep{murphy1996finley} for the list of tokens of the property values associated the same property key in both entities. The Jaccard index involved empty strings is defined as zero in our case. We assign a weight of $0.9$ to the entity name, while all other properties collectively receive a total weight of $0.1$. This ensures that the entity name holds the highest importance for matching, while still acknowledging the contributions of other properties.
It is worthy to notice that the weights values $0.9$ and $0.1$ are not universal standards. One can tailor the choices of these weights values for specific requirements.
Then the optimal assignment matrix $\mathbf{F}$ is found by maximizing the following equation:

\vspace{-5mm}
\begin{small} 
\begin{align}\label{eq:2}
    \mathbf{F}=\argmax_{\mathbf{F}} \sum_{i=1}^{m} \sum_{j=1}^{n} \mathbf{F}_{i,j} \cdot \mathbf{S}_{i,j},
\end{align}
\end{small} %
subject to the following four constraints to ensure one-to-one assignment between entities in the prediction set and the ground truth set: 
\begin{small}
\textit{(\romannumeral1)} $\mathbf{F}_{i,j}\in\{0,1\}$; \textit{(\romannumeral2)} $\sum_{i=1}^{m} \mathbf{F}_{i,j} \leq 1, \forall j \in \{1, 2, \ldots, n\}$; \textit{(\romannumeral3)} $\sum_{j=1}^{n} \mathbf{F}_{i,j} \leq 1, \forall i \in \{1, 2, \ldots, m\}$; \textit{(\romannumeral4)} $\sum_{i=1}^{m}\sum_{j=1}^{n} \mathbf{F}_{i,j} = \min\{m, n\}$. 
\end{small}
One can take an analogy of maximizing equation.~\ref{eq:2} to the optimal flow in the Earth Mover’s Distance (EMD). In EMD, the optimal flow is the one that minimizes the entire ``cost'' of moving the dirt, while in our case, the optimal assignment is the one that maximizes the entire "similarity" in the best possible way.


\vspace{-2mm}
\paragraph{Phase 2: Pairwise Entity Comparison.}
After obtaining the optimal entity assignment, we focus on the pairwise entity comparison.
We define $\psi_\text{ent}(\vec{E'}_{i}, \vec{E}_j)$ as a similarity metric between any two arbitrary entities $e'$ and $e$ from $\mathcal{E}'$ and $\mathcal{E}$.

The pairwise entity similarity function $\psi_\text{ent}$ is defined as a linear average $\bigotimes$ over individual pairwise property similarity $\psi_\text{prop}$ as follows:

\vspace{-5mm}
\begin{small}\label{eq:pairwise}
\begin{align}
    \psi_\text{ent}(e', e) = \bigotimes_{p \in \mathcal{P}} \psi_\text{prop}(v_{e',p}, v_{e,p}), \label{eqn:small-delta-ent}
\end{align}
\end{small} %
where $\psi_\text{prop}(v_{e',p}, v_{e,p})$ is defined as the Jaccard index between the lists of tokens of the predicted values and ground-truth values for corresponding properties.
\new{We define the score as zero for missing properties.}

It should be noted that while both $\mathbf{S}$ and $\psi_{\text{ent}}$ are used to calculate similarities between pairs of entities, they are not identical. During the entity assignment phase, it is more important to make sure the entity names are aligned, while it is more acceptable to treat all properties equally without differentiation during the pairwise entity comparison.
The separation in the definitions of two similarity measures allows us to tailor our metric more precisely to the specific requirements of each phase of the process.
The definition of similarity and different variants for our proposed AESOP metric are elaborated in Appendix~\ref{appendix:diff_variants}. We discuss the relationship between traditional metrics, such as precision and recall, and AESOP in Appendix~\ref{appendix:metric_relation}.





\section{Multi-stage Structured Entity Extraction using Language Models}
\label{sec:model}
\vspace{-2mm}
\begin{figure*}
    \centering
    \includegraphics[width=0.95\linewidth]{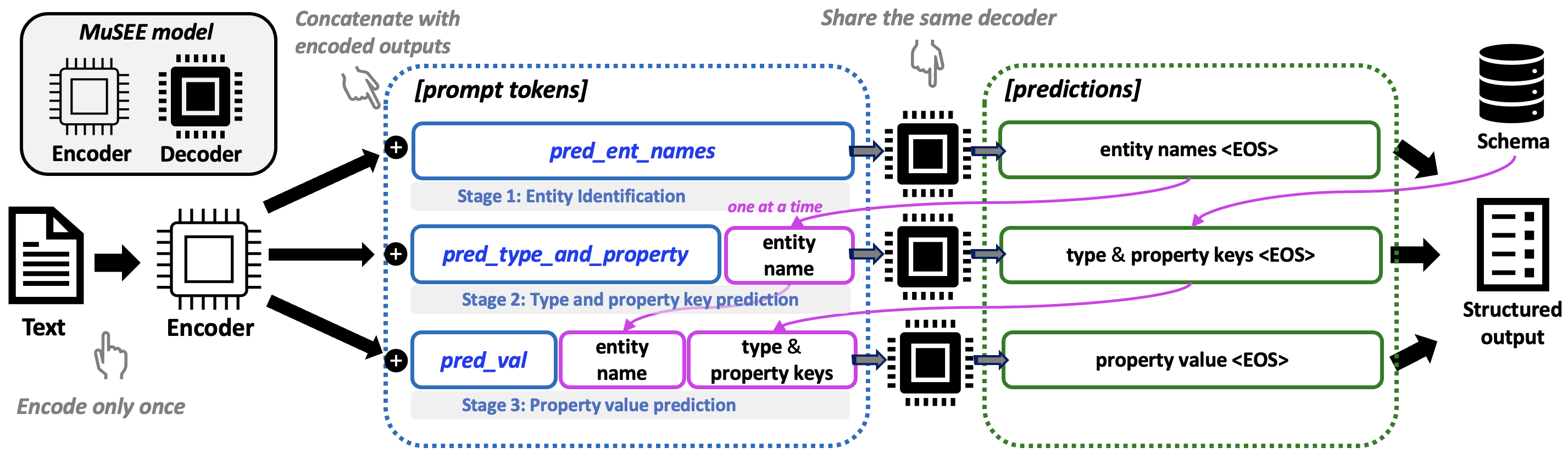}
    \caption{The pipeline of our proposed MuSEE model, which is built on an encoder-decoder architecture. The input text only needs to be encoded once. The decoder is shared for all the three stages. All predictions within each stage can be processed in batch, and teacher forcing enables parallelization even across stages during training.}
    \vspace{-4mm}
    \label{fig:model_architecture}
\end{figure*}
In this section, we elaborate on the methodology for structured entity extraction using LMs.
We introduce a novel model architecture leveraging LMs, \textit{\textbf{MuSEE}}, for \textit{\textbf{Mu}}lti-stage \textit{\textbf{S}}tructured \textit{\textbf{E}}ntity \textit{\textbf{E}}xtaction.
MuSEE is built on an encoder-decoder architecture, whose pipeline incorporates two pivotal enhancements to improve effectiveness and efficiency: \textit{(\romannumeral1)} \textit{reducing output tokens} through introducing additional special tokens where each can be used to replace multiple tokens, and \textit{(\romannumeral2)} \textit{multi-stage parallel generation} for making the model focus on a sub-task at each stage where all predictions within a stage can be processed parallelly.

\vspace{-2mm}
\paragraph{Reducing output tokens.}
Our model condenses the output by translating entity types and property keys into unique, predefined tokens.
Specifically, for the entity type, we add prefix ``\textbf{ent\_type\_}'', while for each property key, we add prefix ``\textbf{pk\_}''. 
By doing so, the type and each property key on an entity is represented by a single token, which significantly reduces the number of output tokens during generation thus improving efficiency. 
For instance, if the original entity type is ``\textit{artificial object}'' which is decomposed into 4 tokens (i.e., ``\_\textit{art}'', ``\textit{if}'', ``\textit{ical}'', ``\_\textit{object}'') using the T5 tokenizer, now we only need one special token, ``\text{\textbf{ent\_type\_}\textit{artifical\_object}}'', to represent the entire sequence.
All of these special tokens can be derived through the knowledge of some predefined schema before the model training.

\vspace{-2mm}
\paragraph{Multi-stage parallel generation.}
In addition to reducing the number of generated tokens, MuSEE further decomposes the generation process into three stages: \textit{(\romannumeral1)} identifying all entities, \textit{(\romannumeral2)} determining entity types and property keys, and \textit{(\romannumeral3)} predicting property values.
To demonstrate this pipeline more clearly, we use the same text shown in Fig.~\ref{fig:task} as an example to show the process of structured entity extraction as follows:

\textbf{\small{Stage 1: Entity Identification.}}
\vspace{-2mm}
\begin{tcolorbox}[width=0.5\textwidth]
\small{
\begin{itemize}[noitemsep, topsep=0pt, partopsep=0pt, parsep=0pt, leftmargin=*]
    \item[\ding{118}] \textcolor{gray}{\pmb{[}Text Description\pmb{]}} $\Rightarrow$ MuSEE $\Rightarrow$ \textcolor{blue}{{\textit{pred\_ent\_names}}}\\ \textbf{``Bill Gates''} \textbf{``Microsoft''} \textbf{$\langle$EOS$\rangle$}
\end{itemize}
}
\end{tcolorbox}

\vspace{-2mm}
\textbf{\small{Stage 2: Type and property key prediction.}}
\vspace{-2mm}
\begin{tcolorbox}[width=0.5\textwidth]
\small{
\begin{itemize}[noitemsep, topsep=0pt, partopsep=0pt, parsep=0pt, leftmargin=*]
    \item[\ding{118}] \textcolor{gray}{\pmb{[}Text Description\pmb{]}} $\Rightarrow$ MuSEE $\Rightarrow$ \textcolor{blue}{{\textit{pred\_type\_and\_property}\\ \{``Bill Gates''\}}} \textbf{ent\_type\_human} \textbf{pk\_country} \textbf{pk\_occupation} \textbf{$\langle$EOS$\rangle$}
    \item[\ding{118}] \textcolor{gray}{\pmb{[}Text Description\pmb{]}} $\Rightarrow$ MuSEE $\Rightarrow$ \textcolor{blue}{{\textit{pred\_type\_and\_property}\\ \{``Microsoft''\}}} \textbf{ent\_type\_corporation} \textbf{pk\_cofounder} \textbf{pk\_headquarter} \textbf{$\langle$EOS$\rangle$}
\end{itemize}
}
\end{tcolorbox}

\vspace{-2mm}
\textbf{\small{Stage 3: Property value prediction.}} 
\vspace{-2mm}
\begin{tcolorbox}[width=0.5\textwidth]
\small{
\begin{itemize}[noitemsep, topsep=0pt, partopsep=0pt, parsep=0pt, leftmargin=*]
    \item[\ding{118}] \textcolor{gray}{\pmb{[}Text Description\pmb{]}} $\Rightarrow$ MuSEE $\Rightarrow$ \textcolor{blue}{\textit{pred\_val}\\ \{``Bill Gates''\} \{ent\_type\_human\} \{pk\_country\}} \textbf{America} \textbf{$\langle$EOS$\rangle$}
    \item[\ding{118}] \textcolor{gray}{\pmb{[}Text Description\pmb{]}} $\Rightarrow$ MuSEE $\Rightarrow$ \textcolor{blue}{\textit{pred\_val}\\ \{``Bill Gates''\} \{ent\_type\_human\} \{pk\_occupation\}} \textbf{Businessman} \textbf{$\langle$EOS$\rangle$}
    \item[\ding{118}] \textcolor{gray}{\pmb{[}Text Description\pmb{]}} $\Rightarrow$ MuSEE $\Rightarrow$ \textcolor{blue}{\textit{pred\_val}\\ \{``Microsoft''\} \{ent\_type\_corporation\} \{pk\_cofounder\}} \textbf{Bill Gates} \textbf{$\langle$EOS$\rangle$}
    \item[\ding{118}] \textcolor{gray}{\pmb{[}Text Description\pmb{]}} $\Rightarrow$ MuSEE $\Rightarrow$ \textcolor{blue}{\textit{pred\_val}\\ \{``Microsoft''\} \{ent\_type\_corporation\} \{pk\_headquarter\}} \textbf{Redmond} \textbf{$\langle$EOS$\rangle$}
\end{itemize}
}
\end{tcolorbox}

Among the three stages depicted, \textit{\textcolor{blue}{pred\_ent\_names}}, \textit{\textcolor{blue}{pred\_type\_and\_property}}, and \textit{\textcolor{blue}{pred\_val}} are special tokens to indicate the task.
For each model prediction behavior, the first ``$\Rightarrow$'' indicates inputting the text into the encoder of MuSEE, while the second ``$\Rightarrow$'' means inputting the encoded outputs into the decoder.
All tokens in \textcolor{blue}{blue} are the prompt tokens input into the decoder which do not need to be predicted, while all tokens in \textbf{bold} are the model predictions.
For the stage 1, we emphasize that MuSEE outputs a unique identifier for each entity in the given text. Taking the example in Fig.~\ref{fig:task}, the first stage outputs ``\textit{Bill Gates}'' only, rather than both ``\textit{Bill Gates}'' and ``\textit{Gates}''. This requires the model implicitly learn how to do coreference resolution, namely learning that ``\textit{Bill Gates}'' and ``\textit{Gates}'' are referring to the same entity. Therefore, our approach uses neither surface forms, as the outputs of the first stage are unique identifiers, nor the entity titles followed by entity linkings.
For stage 2, the MuSEE model predicts the entity types and property keys, which are all represented by special tokens. Hence, the prediction can be made by sampling the token with highest probability over the special tokens for entity types and property keys only, rather than all tokens.
Notice that we do not need to predict the value for ``\textit{type}'' and ``\textit{name}'' in stage 3, since the type can be directly derived from the ``\textbf{ent\_type\_}'' special key itself, and the name is obtained during stage 1.
The tokens in the bracket ``\{..\}'' are also part of the prompt tokens and are obtained in different ways during training and inference.
During training, these inputs are obtained from the ground truth due to the teacher forcing technique~\citep{raffel2023exploring}.
During inference, they are obtained from the output predictions from the previous stages.
The full training loss is a sum of three cross-entropy losses, one for each stage.
An illustration of our model's pipeline is shown in Fig.~\ref{fig:model_architecture}.
More implementation details are elaborated in Appendix~\ref{appendix:details}.

\vspace{-3mm}
\paragraph{Benefits for Training and Inference.}
MuSEE's unique design benefits both training and inference. 
In particular, each stage in MuSEE is finely tuned to concentrate on a specific facet of the extraction process, thereby enhancing the overall effectiveness.
Most importantly, all predictions within the same stage can be processed in batch thus largely improving efficiency.
The adoption of a teacher forcing strategy enables parallel training even across different stages, further enhancing training efficiency. 
During inference, the model’s approach to breaking down long sequences into shorter segments significantly reduces the generation time.
It is also worthy to mention that each text in the above three stages needs to be encoded only once by the MuSEE's encoder, where the encoded output is repeatedly utilized across different stages.
This streamlined approach ensures a concise and clear delineation of entity information, facilitating the transformation of unstructured text into a manageable and structured format.

\section{Experiments}
\label{sec:experiment}
\vspace{-2mm}
In this section, we describe the datasets used in our experiment, followed by the discussion of baseline methods and training details. 
\subsection{Data}
\label{sec:experiment-data}
\vspace{-2mm}
In adapting the structured entity extraction, we repurpose the NYT~\cite{NYT}, CoNLL$04$~\cite{conll04}, and REBEL~\cite{huguet-cabot-navigli-2021-rebel-relation} datasets, which are originally developed for relation extractions.
For NYT and CoNLL$04$, since each entity in these two datasets has a predefined type, we simply reformat them to our entity-centric formulation by treating the subjects as entities, relations as property keys, and objects as property values. 
%
%
REBEL connects entities identified in Wikipedia abstracts as hyperlinks, along with dates and values, to entities in Wikidata and extracts the relations among them.
%
%
For entities without types in the REBEL dataset, we categorize their types as ``\textit{unknown}''.
Additionally, we introduce a new dataset, named Wikidata-based.
The Wikidata-based dataset is crafted using an approach similar to REBEL but with two primary distinctions: 
\textit{(\romannumeral1)} property values are not necessarily entities;
\textit{(\romannumeral2)} we simplify the entity types by consolidating them into broader categories based on the Wikidata taxonomy graph, resulting in less specific types.
%
%
%
The processes for developing the Wikidata-based dataset is detailed in Appendix~\ref{appendix:dataset}.
The predefined schemas for NYT, CoNLL$04$, and REBEL are using all entity types and property keys from these datasets. 
The details of the predefined schema for Wikidata-based dataset are provided in Appendix~\ref{appendix:dataset}.
Comprehensive statistics for all four datasets are available in Appendix~\ref{appendix:dataset_stats}.






\vspace{-2mm}
\subsection{Baseline}
\label{sec:experiment-baseline}
\vspace{-2mm}
We benchmark our methodology against two distinct classes of baseline approaches.
%
%
The first category considers adaptations from general seq2seq task models:
\textit{(\romannumeral1)} LM-JSON: this approach involves fine-tuning pre-trained language models. 
The input is a textual description, and the output is the string format JSON containing all entities.
The second category includes techniques designed for different information extraction tasks, which we adapt to address our challenge:
\textit{(\romannumeral2)} GEN2OIE~\cite{kolluru-etal-2022-alignment}, which employs a two-stage generative model initially outputs relations for each sentence, followed by all extractions in the subsequent stage;
\textit{(\romannumeral3)} IMoJIE~\cite{kolluru2020imojie}, an extension of CopyAttention~\cite{cui2018neural}, which sequentially generates new extractions based on previously extracted tuples; 
\textit{(\romannumeral4)} GenIE~\cite{josifoski2022genie}, an end-to-end autoregressive generative model using a bi-level constrained generation strategy to produce triplets that align with a predefined schema for relations. 
GenIE is crafted for the closed information extraction, so it includes a entity linking step. However, in our strict setting, there is only a schema of entity types and relations. Therefore, we repurpose GenIE for our setting by maintaining the constrained generation strategy and omitting the entity linking step.
We omit to compare our method with non-generative models primarily due to the task differences.
%

\vspace{-2mm}
\subsection{Training}
\label{sec:experiment-training}
\vspace{-2mm}
%
%
We follow existing studies~\cite{huguet-cabot-navigli-2021-rebel-relation} to use the encoder-decoder architecture in our experiment. 
%
We choose the T5~\cite{raffel2023exploring} series of LMs and employ the pre-trained T5-Base (T5-B) and T5-Large (T5-L) as the base models underlying every method discussed in section~\ref{sec:experiment-baseline} and our proposed MuSEE.
LM-JSON and MuSEE are trained with the Low-Rank Adaptation~\cite{hu2021lora}, where $r=16$ and $\alpha=32$.
%
%
For GEN2OIE, IMoJIE, and GenIE, we follow all training details of their original implementation.
For all methods, we employ a linear warm up and the Adam optimizer~\cite{kingma2017adam}, tuning the learning rates between 3$e$-4 and 1$e$-4, and weight decays between 1$e$-2 and 0.
%
All experiments are run on a NVIDIA A$100$ GPU.

It is worthy to mention that MuSEE can also build upon the decoder-only architecture by managing the KV cache and modifications to the position encodings~\cite{xiao2024efficient}, though this requires additional management and is not the main focus of this study.



\vspace{-3mm}
\section{Results}
\label{sec:result}
\vspace{-2mm}
In this section, we show the results for both quantitative and human side-by-side evaluation.

\begin{table*}[t]
    \centering
    \captionsetup{width=1.0\linewidth}
    \caption{Summary of results of different models. Each metric is shown in percentage (\%). The last column shows the inference efficiency, measured by the number of samples the model can process per second. The best is \textbf{bolded}, and the second best is \underline{underlined}.
    Our model has a statistical significance for $p\leq 0.01$ compared to the best baseline (labelled with *) based on the paired t-test.
    }
    \label{tab:multiprop}
    \resizebox{1.0\textwidth}{!}
    {
    \begin{tabular}{l|cccc|cccc|cccc|cccc|c}
    \toprule 
        \multirow{2}{*}{Model} &\multicolumn{4}{c|}{\textbf{REBEL}} &\multicolumn{4}{c|}{\textbf{NYT}}  &\multicolumn{4}{c|}{\textbf{CoNLL04}} &\multicolumn{4}{c|}{\textbf{Wikidata-based}} & \multirow{2}{*}{\makecell{samples \\per sec}}\\
        & AESOP & Precision & Recall & F1 & AESOP & Precision & Recall & F1 & AESOP & Precision & Recall & F1 & AESOP & Precision & Recall & F1\\
        \midrule
         LM-JSON \small{(T5-B}) & 41.91 & 38.33 & \textbf{51.29} & 43.87 & 66.33 & 73.10 & 52.66 & 61.22 & 68.80 & 61.63 & 48.04 & 53.99 & 36.98 & 43.95 & 29.82 & 35.53 & 19.08 \\
         GEN2OIE \small{(T5-B}) & 44.52 & 35.23 & 40.28 & 37.56 & 67.04 & 72.08 & 53.02 & 61.14 & 68.39 & 62.35 & 42.20 & 50.26 & 37.07 & 40.87 & 28.37 & 33.55 & \underline{28.21}\\
         IMoJIE \small{(T5-B}) & 46.11 & 34.10 & \underline{48.61} & 40.08 & 63.86 & 72.28 & 48.99 & 58.40 & 63.68 & 52.00 & 42.62 & 46.85 & 37.08 & 41.61 & 28.23 & 33.64 & 5.36\\
         GenIE \small{(T5-B}) & \underline{48.82}$^*$ & \textbf{57.55} & 38.70 & \underline{46.28}$^*$ & \underline{79.41}$^*$ & \underline{87.68} & \textbf{73.24} & \textbf{79.81} & \underline{74.74}$^*$ & \underline{72.49}$^*$ & \underline{59.39} & \underline{65.29} & \underline{40.60}$^*$ & \underline{50.27}$^*$ & \textbf{29.75} & \underline{37.38} & 10.19 \\
         \textbf{MuSEE \small{(T5-B)}} & \textbf{55.24} & \underline{56.93} & 42.31 & \textbf{48.54} & \textbf{81.33} & \textbf{88.29} & \underline{72.21} & \underline{79.44} & \textbf{78.38} & \textbf{73.18} & \textbf{60.28} & \textbf{66.01} & \textbf{46.95} & \textbf{53.27} & \underline{29.33} & \textbf{37.99} &  \textbf{52.93}\\
         \midrule\midrule
         LM-JSON \small{(T5-L)} & 45.92 & 39.49 & 40.82 & 40.14 & 67.73 & 73.38 & 53.22 & 61.69 & 68.88 & 61.50 & 47.77 & 53.77 & 38.19 & 43.24 & 31.63 & 36.54 & 11.24\\
         GEN2OIE \small{(T5-L}) & 46.70 & 37.28 & 41.12 & 39.09 & 68.27 & 73.97 & 53.32 & 61.88 & 68.52  & 62.76 & 43.31 & 51.16 & 38.25 & 41.23 & 28.54 & 33.77 & \underline{18.56} \\
         IMoJIE \small{(T5-L}) & 48.13 & 38.55 & \textbf{49.73} & 43.43 &65.72 & 73.46 & 50.03 & 59.52 & 67.31 & 53.00 & 43.44 & 47.75 & 38.18 & 41.74 & 30.10 & 34.98 & 3.73\\
         GenIE \small{(T5-L)} & \underline{50.06}$^*$ & \textbf{58.00} & 42.56 & \textbf{49.09} & \underline{79.64}$^*$ & \underline{84.82}$^*$ & \textbf{75.69} & \underline{80.00} & \underline{72.92}$^*$ & \textbf{77.75} & \underline{55.64}$^*$ &\underline{64.86} & \underline{43.50}$^*$ & \textbf{54.05} & \underline{30.98} & \textbf{39.38} & 5.09\\
         \textbf{MuSEE \small{(T5-L)}} & \textbf{57.39} & \underline{57.11} & \underline{42.89} & \underline{48.96} & \textbf{82.67} & \textbf{89.43} & \underline{73.32} & \textbf{80.60} & \textbf{79.87} & \underline{74.89} & \textbf{60.72} & \textbf{67.08} & \textbf{50.94} & \underline{53.72} & \textbf{31.12} & \underline{39.24} &  \textbf{33.96} \\

         \bottomrule
    \end{tabular}
    }
    \vspace{-4mm}
\end{table*}

\vspace{-3mm}
\subsection{Quantitative Evaluation}

\paragraph{Effectiveness comparison.} 
The overall effectiveness comparison is shown in Table~\ref{tab:multiprop}.
We report traditional metrics, including precision, recall, and F1 score, in addition to our proposed AESOP metric.
%
From the results, the MuSEE model consistently outperforms other baselines in terms of AESOP across all datasets.
%
For instance, MuSEE achieves the highest AESOP scores on REBEL with 55.24 (T5-B) and 57.39 (T5-L), on NYT with 81.33 (T5-B) and 82.67 (T5-L), on CoNLL04 with 78.38 (T5-B) and 79.87 (T5-L), and on the Wikidata-based dataset with 46.95 (T5-B) and 50.94 (T5-L).
These scores significantly surpass those of the competing models, indicating MuSEE's stronger entity extraction capability.
The other three traditional metrics further underscore the efficacy of the MuSEE model.
For instance, on CoNLL04, MuSEE (T5-B) achieves a precision of 73.18, a recall of 60.28, and a F1 score of 66.01, which surpass all the other baselines.
Similar improvements are observed on REBEL, NYT, and Wikidata-based dataset.
Nevertheless, while MuSEE consistently excels in the AESOP metric, it does not invariably surpass the baselines across all the traditional metrics of precision, recall, and F1 score. Specifically, within the REBEL dataset, GenIE (T5-B) achieves the highest precision at 57.55, and LM-JSON (T5-B) records the best recall at 51.29. Furthermore, on the NYT dataset, GenIE (T5-B) outperforms other models in F1 score. These variances highlight the unique insights provided by our adaptive AESOP metric, which benefits from our entity-centric formulation. We expand on this discussion in section~\ref{sec:human_eval}.

As discussed in Sec.~\ref{sec:model}, our MuSEE model is centered around two main enhancements: reducing output tokens and multi-stage parallel generation.
By simplifying output sequences, MuSEE tackles the challenge of managing long sequences that often hinder baseline models, like LM-JSON, GenIE, IMoJIE, thus reducing errors associated with sequence length.
Additionally, by breaking down the extraction process into three focused stages, MuSEE efficiently processes each aspect of entity extraction, leveraging contextual clues for more accurate predictions. 
In contrast, GEN2OIE's two-stage approach, though similar, falls short because it extracts relations first and then attempts to pair entities with these relations.
However, a single relation may exist among different pairs of entities, which can lead to low performance with this approach.
Supplemental ablation study is provided in Appendix~\ref{appendix:ablation}.
\vspace{-3mm}
\paragraph{Efficiency comparison.}
As shown in the last column of Table~\ref{tab:multiprop}, we provide a comparison on the inference efficiency, measured in the number of samples the model can process per second.
The MuSEE model outperforms all baseline models in terms of efficiency, processing 52.93 samples per second with T5-B and 33.96 samples per second with T5-L. 
It shows a 10x speed up compared to IMoJIE, and a 5x speed up compared to the strongest baseline GenIE.
This high efficiency can be attributed to MuSEE's architecture, specifically its multi-stage parallel generation feature. 
By breaking down the task into parallelizable stages, MuSEE minimizes computational overhead, allowing for faster processing of each sample.
The benefit of this design can also be approved by the observation that the other multi-stage model, GEN2OIE, shows the second highest efficiency.

\begin{figure}[t]
    \centering
    \includegraphics[width=\linewidth]{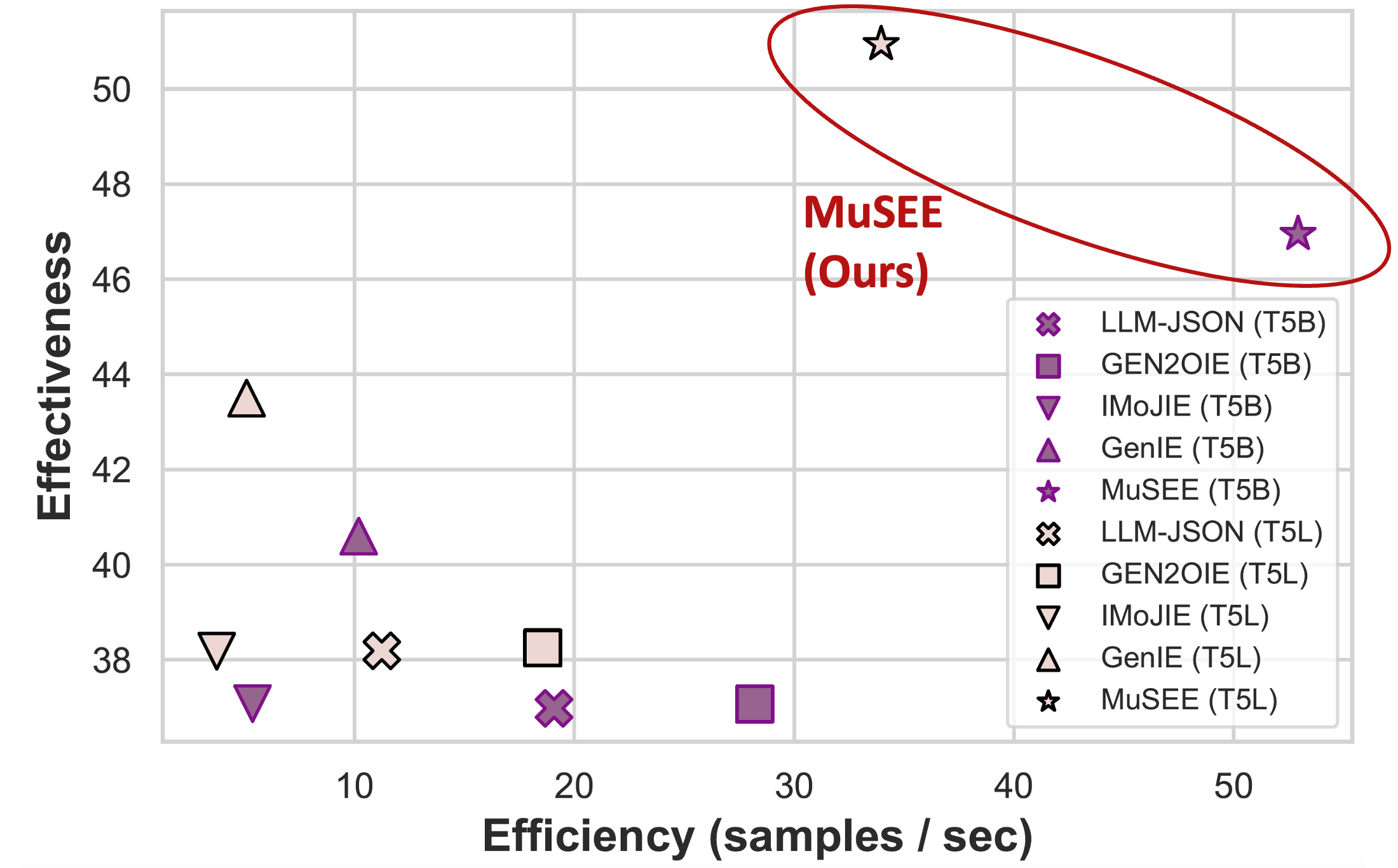}
    \caption{An overall effectiveness-and-efficiency comparison across models on Wikidata-based Dataset. MuSEE strongly outperforms all baselines on both measures. The effectiveness is measured by AESOP.}
    \vspace{-4mm}
    \label{fig:ee_tradeoff}
\end{figure}

\vspace{-1mm}
To better illustrate our model's strength, we show the scatter plots  comparing all models with various backbones in Fig.~\ref{fig:ee_tradeoff} on the effectiveness and efficiency.
We choose the Wikidata-based dataset and the effectiveness is measured by AESOP.
As depicted, our model outperforms all baselines with a large margin.
This advantage makes MuSEE particularly suitable for applications requiring rapid processing of large volumes of data, such as processing web-scale datasets, or integrating into interactive systems where response time is critical.

\begin{figure}[t]
    \centering
    \includegraphics[width=\linewidth]{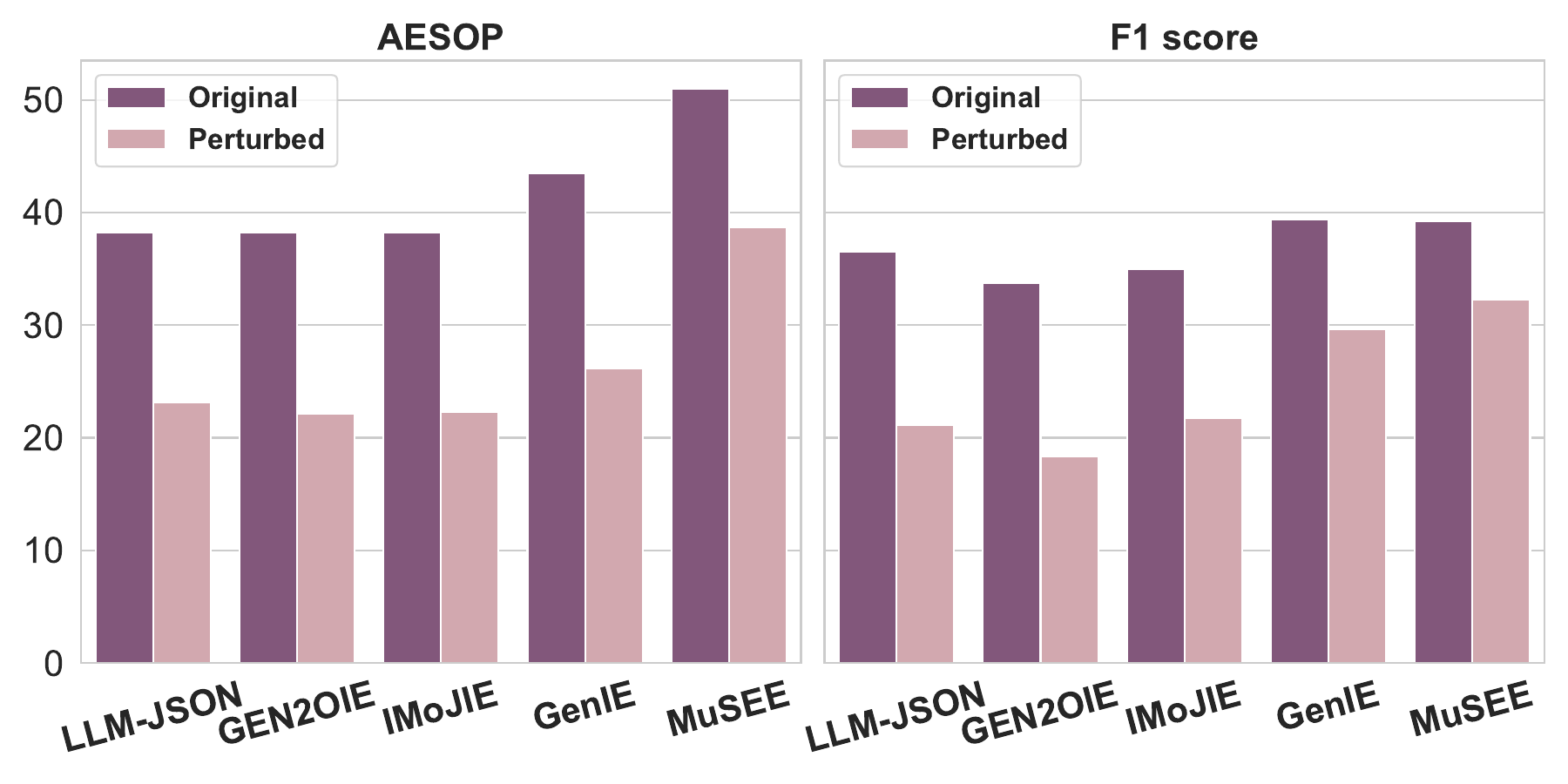}
    \caption{Grounding check across models on the Wikidata-based dataset. MuSEE shows the least performance drop on the perturbed version of data compared to other baselines.}
    \vspace{-6mm}
    \label{fig:hallucination}
\end{figure}


\vspace{-3mm}
\paragraph{Grounding check.}
As the family of T5 models are pre-trained on Wikipedia corpus~\cite{raffel2023exploring}, we are curious whether the models are extracting information from the given texts, or they are leveraging their prior knowledge to generate information that cannot be grounded to the given description.
We use T5-L as the backbone in this experiment. 
We develop a simple approach to conduct this grounding check by perturbing the original test dataset with the following strategy.
We first systematically extract and categorize all entities and their respective properties, based on their entity types.
Then, we generate a perturbed version of the dataset, by randomly modifying entity properties based on the categorization we built.
We introduce controlled perturbations into the dataset by selecting alternative property values from the same category but different entities, and subsequently replacing the original values in the texts.
The experiment results from our grounding study on the Wikidata-based dataset, as illustrated in Fig.~\ref{fig:hallucination}, reveal findings regarding the performance of various models under the AESOP and F1 score.
Our model, MuSEE, shows the smallest performance gap between the perturbed data and the original data compared to its counterparts, suggesting its stronger capability to understand and extract structured information from given texts.
%

\vspace{-2mm}
\subsection{Human Evaluation} \label{sec:human_eval}
\vspace{-2mm}
To further analyze our approach, we randomly select 400 test passages from the Wikidata-based dataset, and generate outputs of our model MuSEE and the strongest baseline GenIE. 
Human evaluators are presented with a passage and two randomly flipped extracted sets of entities with properties. 
Evaluators are then prompted to choose the output they prefer or express no preference based on three criteria, \textit{Completeness}, \textit{Correctness}, and \textit{Hallucinations} (details shown in Appendix~\ref{appendix:human_eval}).
Among all 400 passages, the output of MuSEE is preferred 61.75\% on the completeness, 59.32\% on the correctness, and 57.13\% on the hallucinations.
For a complete comparison, we also report the percentage of samples preferred by quantitative metrics on MuSEE's results when compared with GenIE's, as summarized in Table~\ref{tab:compare}.
As shown, our proposed AESOP metric aligns more closely with human judgment than traditional metrics.
These observations provide additional confirm to the quantitative results evaluated using the AESOP metric that our model significantly outperforms existing baselines and illustrates the inadequacy of traditional metrics due to their oversimplified assessment of extraction quality.
Case study of the human evaluation is shown in Appendix~\ref{appendix:human_eval}.

\begin{table}[t]
    \centering
    \resizebox{0.48\textwidth}{!}
    {
    \begin{tabular}{c|ccc|cccc}
    \toprule
         & \multicolumn{3}{c|}{Human Evaluation} & \multicolumn{4}{c}{Quantitative Metrics} \\
         & Complete. & Correct. & Halluc. & AESOP & Precision & Recall & F1\\
         \midrule
        \makecell{MuSEE prefer} & 61.75 & 59.32 & 57.13 & 61.28 & 45.33 & 37.24 & 40.57\\
        \bottomrule
    \end{tabular}
    }
    \vspace{1mm}
    \caption{Percentage of samples preferred by humans and metrics on MuSEE's results when compared with GenIE's. The first three columns are for human evaluation. The next four columns are for quantitative metrics.}
    \label{tab:compare}
    \vspace{-6mm}
\end{table}
\vspace{-2mm}
\section{Discussion and Conclusion}
\label{sec:conclusion}
\vspace{-2mm}
We introduce Structured Entity Extraction (SEE), an entity-centric formulation of information extraction in a strict setting.
We then propose the Approximate Entity Set OverlaP (AESOP) Metric, which focuses on the entity-level and more flexible to include different level of normalization.
Based upon, we propose a novel model architecture, MuSEE, that enhances both effectiveness and efficiency.
Both quantitative evaluation and human side-by-side evaluation confirm that our model outperforms baselines.

An additional advantage of our formulation is its potential to address coreference resolution challenges, particularly in scenarios where multiple entities share the same name or lack primary identifiers such as names. 
Models trained with prior triplet-centric formulation cannot solve the above challenges.
However, due to a scarcity of relevant data, we were unable to assess this aspect in our current study. 

\vspace{-2mm}
\section{Limitations}
\vspace{-2mm}
The limitation of our work lies in the assumption that each property possesses a single value. 
However, there are instances where a property's value might consist of a set, such as varying ``names''.
Adapting our method to accommodate these scenarios presents a promising research direction.

\section{Acknowledgement}
We would like to thank all reviewers for their professional review work, constructive comments, and valuable suggestions on our manuscript. This work is supported by the the MSR-Mila Research Grant. We thank Compute
Canada for the computing resources.

\bibliography{custom}
\bibliographystyle{acl_natbib}

\appendix
\onecolumn
\section{Variants of AESOP} \label{appendix:diff_variants}

The AESOP metric detailed in section~\ref{sec:aesop} matches entities by considering all properties and normalizes with the maximum of the sizes of the target set and the predicted set. We denote it as AESOP-MultiProp-Max.
In this section, we elaborate more variants of the AESOP metric in addition to section~\ref{sec:aesop}, categorized based on two criteria: the definition of entity similarity used for entity assignment and the normalization approach \new{when computing the final metric value between $\mathcal{E}'$ and $\mathcal{E}$}.
These variants allow for flexibility and adaptability to different scenarios and requirements in structured entity extraction.

\paragraph{Variants Based on Entity Assignment.}
The first category of variants is based on the criteria for matching entities between the prediction $\mathcal{E}'$ and the ground-truth $\mathcal{E}$. 
We define three variants:

\begin{itemize}[wide=0pt, leftmargin=*]
\item \textbf{AESOP-ExactName}: Two entities are considered a match if their names are identical, disregarding case sensitivity.
    This variant is defined as $\mathbf{S}_{i,j}=1$ if $ v_{e'_i, \text{name}} = v_{e_j, \text{name}}$, otherwise 0.
\item \textbf{AESOP-ApproxName}: Entities are matched based on the similarity of their ``\textit{name}'' property values. This similarity can be measured using a text similarity metric, such as the Jaccard index.
\item \textbf{AESOP-MultiProp}: Entities are matched based on the similarity of all their properties, with a much higher weight given to the ``\textit{entity name}'' property due to its higher importance.
\end{itemize}

\paragraph{Variants Based on Normalization.}
The second category of variants involves different normalization approaches \new{for computing the final metric value through Eq.~\ref{eqn:big-delta}}:
\begin{itemize}[wide=0pt, leftmargin=*]
\item \textbf{AESOP-Precision}: The denominator is the size of the predicted set $\mathcal{E}'$, i.e., $\mu=m$.
\item \textbf{AESOP-Recall}: The denominator is the size of the target set $\mathcal{E}$, i.e., $\mu=n$.
\item \textbf{AESOP-Max}: The denominator is the maximum of the sizes of the target set and the predicted set, i.e., $\mu=\max\{m, n\}$.
\end{itemize}
Given these choices, we can obtain $3\times3=9$ variants of the AESOP metric.
To avoid excessive complexity, we regard the AESOP-MultiProp-Max as default.
For clarity, we illustrate the two phases of computing the AESOP metric and its variants in Fig.~\ref{fig:metric}.
We also show that precision and recall are specific instances of the AESOP metric in Appendix~\ref{appendix:metric_relation}.

\begin{figure}[h]
    \centering
    \includegraphics[width=\linewidth]{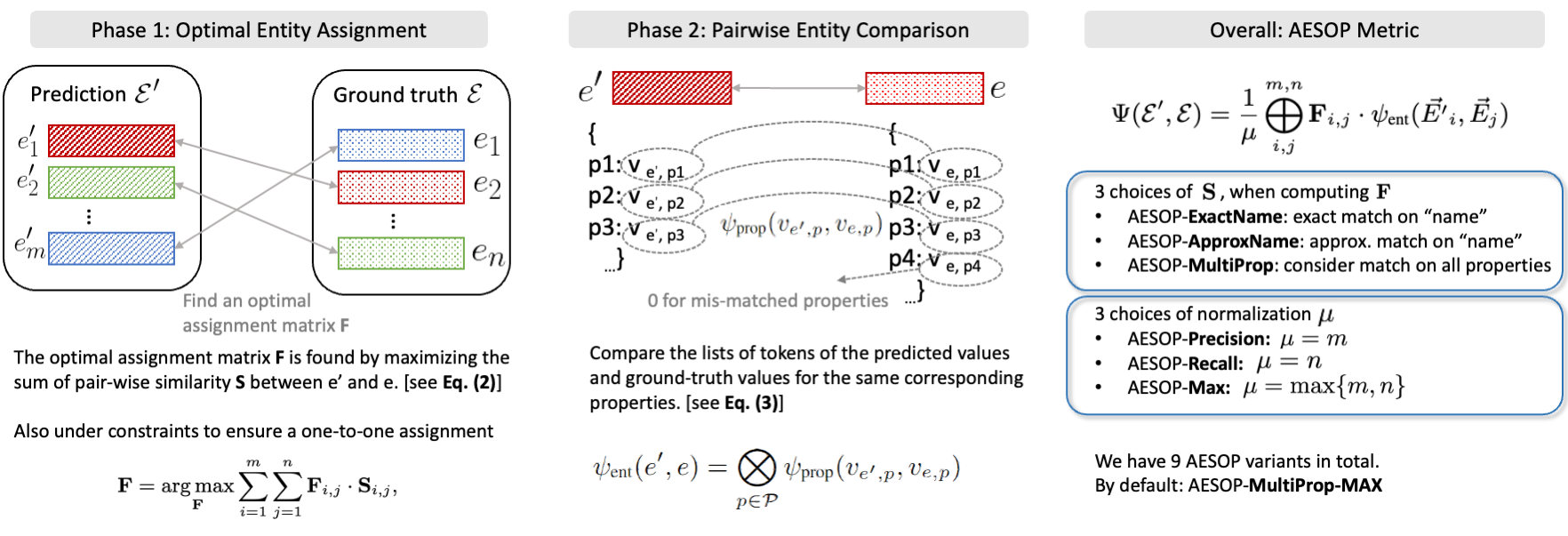}
    \caption{An illustration of the AESOP metric, including optimal entity assignment (phase 1) and pairwise entity comparison (phase 2), and overall metric computation with various similarity and normalization  choices.}
    \label{fig:metric}
\end{figure}

\section{Relationship between Precision/Recall and AESOP}
\label{appendix:metric_relation}
In this section, we show the traditional metrics, precision and recall, are specific instances of the AESOP metric.
To calculate precision and recall, we use the following equations on the number of triplets, where each triplet contains \textit{subject}, \textit{relation}, and \textit{object}.
\begin{equation}
    \text{precision} = \frac{\text{\# of correctly predicted triplets}}{\text{\# of triplets in the prediction}},
\end{equation}
\begin{equation}
    \text{recall} = \frac{\text{\# of correctly predicted triplets}}{\text{\# of triplets in the target}}.
\end{equation}
In the framework of the AESOP metric, precision and recall are effectively equivalent to treating each triplet as an entity, where the \textit{subject} as the entity name, and the \textit{relation} and \textit{object} form a pair of property key and value.
For optimal entity assignment (phase 1), precision and recall use the AESOP-MultiProp variant but match entities based on the similarity of all their properties with a same weight.
For pairwise entity comparison (phase 2), the $\psi_\text{ent}(e', e)$ (Eq.~\ref{eqn:small-delta-ent}), can be defined as 1 if $v'=v$, otherwise 0,
where $v'$ and $v$ are the only property values in $e'$ and $e$, respectively.
%
For Eq.~\ref{eqn:big-delta}, $\bigoplus$ aggregation can be defined as a linear sum, which principally results in how many triplets are correctly predicted in this case.
If $\mu$ in Eq.~\ref{eqn:big-delta} is set as the number of triplets in the prediction, this corresponds to the calculation of precision. 
Similarly, when $\mu$ equals the number of triplets in the target, it corresponds to the calculation of recall.

\section{Implementation Details of MuSEE} \label{appendix:details}
In order to implement the approach of our MuSEE model, one may extend existing models with encoder-decoder architecture by integrating additional modules and processing steps specifically designed for entity and property prediction tasks. 
Specifically, given a predefined schema, we first add all necessary special tokens to customize the tokenizer as detailed before. 
The implementation of the generation process involves three main stages: entity prediction, property key prediction, and property value prediction.

\begin{enumerate}
    \item Entity Prediction: We first encode the input sequence using the encoder to obtain the hidden states for the entire sequence. We generate a prompt ``pred\_ent\_names'' and transform it to token ids using the tokenizer. This prompt, repeated for each sample in the batch, is concatenated with the encoded input sequence and processed through the decoder to produce entity name predictions as a sequence of tokens.
    \item Property Key Prediction: For each predicted entity name, we generate prompts in the format ``pred\_type\_and\_property [entity\_name]''. These prompts are tokenized, padded to a fixed length, and concatenated with the encoded input sequence. The concatenated sequences are then passed through the decoder to predict entity types and property keys as a sequence of special tokens for entity types and property keys. We achieve this by sampling the token with highest probability over all special tokens for entity types and property keys, rather than training a separate classifier head.
    \item Property Value Prediction: For each predicted entity and its corresponding property keys, we create prompts in the format ``pred\_val [entity\_name] [entity\_type] [property\_key]''. These prompts are tokenized, padded, and concatenated with the encoded input sequence. The concatenated sequences are processed by the decoder to generate property value predictions.
\end{enumerate}

The training loss is a summation of the cross-entropy loss from each stage, and the training process can be parallel as we elaborate in section~\ref{sec:model}.


\section{Details of Wikidata-based Dataset} \label{appendix:dataset}
We build a new Wikidata-based dataset.
This dataset is inspired by methodologies employed in previous works such as Wiki-NRE~\cite{trisedya-etal-2019-neural}, T-REx~\cite{elsahar-etal-2018-rex}, REBEL~\cite{huguet-cabot-navigli-2021-rebel-relation}, leveraging extensive information available on Wikipedia and Wikidata.
The primary objective centers around establishing systematic alignments between textual content in Wikipedia articles, hyperlinks embedded within these articles, and their associated entities and properties as cataloged in Wikidata. 
This procedure is divided into three steps:
\textit{(i) Parsing Articles:} 
We commence by parsing English Wikipedia articles from the dump file\footnote{The version of the Wikipedia and Wikidata dump files utilized in our study are 20230720, representing the most recent version available during the development of our work.\label{note1}}, focusing specifically on text descriptions and omitting disambiguation and redirect pages. 
The text from each selected article is purified of Wiki markup to extract plain text, and hyperlinks within these articles are identified as associated entities. 
Subsequently, the text descriptions are truncated to the initial ten sentences, with entity selection confined to those referenced within this truncated text. 
This approach ensures a more concentrated and manageable dataset.
\textit{(ii) Mapping Wikidata IDs to English Labels:} 
Concurrently, we process the Wikidata dump\footnotemark[1] file to establish a mapping (termed as the \textit{id-label map}) between Wikidata IDs and their corresponding English labels. 
This mapping allows for efficient translation of Wikidata IDs to their English equivalents.
%
\textit{(iii) Interconnecting Wikipedia articles with Wikidata properties:} 
For each associated entity within the text descriptions, we utilize Wikidatas API to ascertain its properties and retrieve their respective Wikidata IDs. 
The previously established \textit{id-label map} is then employed to convert these property IDs into English labels. 
Each entitys type is determined using the value associated with \textit{instance of (P31)}. 
Given the highly specific nature of these entity types (e.g., \textit{small city (Q18466176)}, \textit{town (Q3957)}, \textit{big city (Q1549591)}), we implement a recursive merging process to generalize these types into broader categories, referencing the \textit{subclass of (P279)} property. 
Specifically, we first construct a hierarchical taxonomy graph.
Each node within this graph structure represents an entity type, annotated with a count reflecting the total number of entities it encompasses.
Second, a priority queue are utilized, where nodes are sorted in descending order based on their entity count. 
We determine whether the top $n$ nodes represent an ideal set of entity types, ensuring the resulted entity types are not extremely specific. 
Two key metrics are considered for this evaluation: the percentage of total entities encompassed by the top $n$ nodes, and the skewness of the distribution of each entity type's counts within the top $n$ nodes.
If the distribution is skew, we then execute a procedure of dequeuing the top node and enqueueing its child nodes back into the priority queue. 
This iterative process allows for a dynamic exploration of the taxonomy, ensuring that the most representative nodes are always at the forefront.
%
Finally, our Wikidata-based dataset is refined to contain the top-10 (i.e., $n=10$) most prevalent entity types according to our hierarchical taxonomy graph and top-10 property keys in terms of occurrence frequency, excluding entity name and type.
The 10 entity types are \textit{talk, system, spatio-temporal entity, product, natural object, human, geographical feature, corporate body, concrete object,} and \textit{artificial object}.
The 10 property keys are \textit{capital, family name, place of death, part of, location, country, given name, languages spoken, written or signed, occupation,} and \textit{named after}.

\section{Statistics of Datasets}\label{appendix:dataset_stats}
NYT is under the CC-BY-SA license. CoNLL$04$ is under the Creative Commons Attribution-NonCommercial-ShareAlike $3.0$ International License. REBEL is under the Creative Commons Attribution $4.0$ International License.
The dataset statistics presented in Table~\ref{tab:dataset_stats} compare NYT, CoNLL$04$, REBEL, and Wikidata-based datasets. 
All datasets feature a minimum of one entity per sample, but they differ in their average and maximum number of entities, with the Wikidata-based dataset showing a higher mean of $3.84$ entities.
They also differ in the maximum number of entities, where REBEL has a max of $65$. 
Property counts also vary, with REBEL having a slightly higher average number of properties per entity at $3.40$. 
%


\begin{table*}[ht]
    \centering
    \caption{Statistics of all three datasets used in our paper.}
    \label{tab:dataset_stats}
    {
    \begin{tabular}{l|cccc}
    \toprule 
        \multirow{1}{*}{Statistics} & NYT & CoNLL$04$ & REBEL & Wikidata-based \\
        \midrule
         \# of Entity Min & 1 & 1 & 1 & 1\\
         \# of Entity Mean & 1.25 & 1.22 & 2.37 & 3.84\\
         \# of Entity Max & 12 & 5 & 65 & 20\\
         \# of Property Min & 3 & 3 & 2 & 2\\
         \# of Property Mean & 3.19 & 3.02 & 3.40 & 2.80\\
         \# of Property Max & 6 & 4 & 17 & 8\\
         \# of Training Samples & 56,196 & 922 & 2,000,000 & 23,477\\
         \# of Testing Samples & 5,000 & 288 & 5,000 & 4,947\\
         
         \bottomrule
    \end{tabular}
    }
\end{table*}

\begin{figure}[ht]
\centering
\begin{minipage}[t]{.49\textwidth}
  \centering
    \includegraphics[width=0.9\columnwidth]{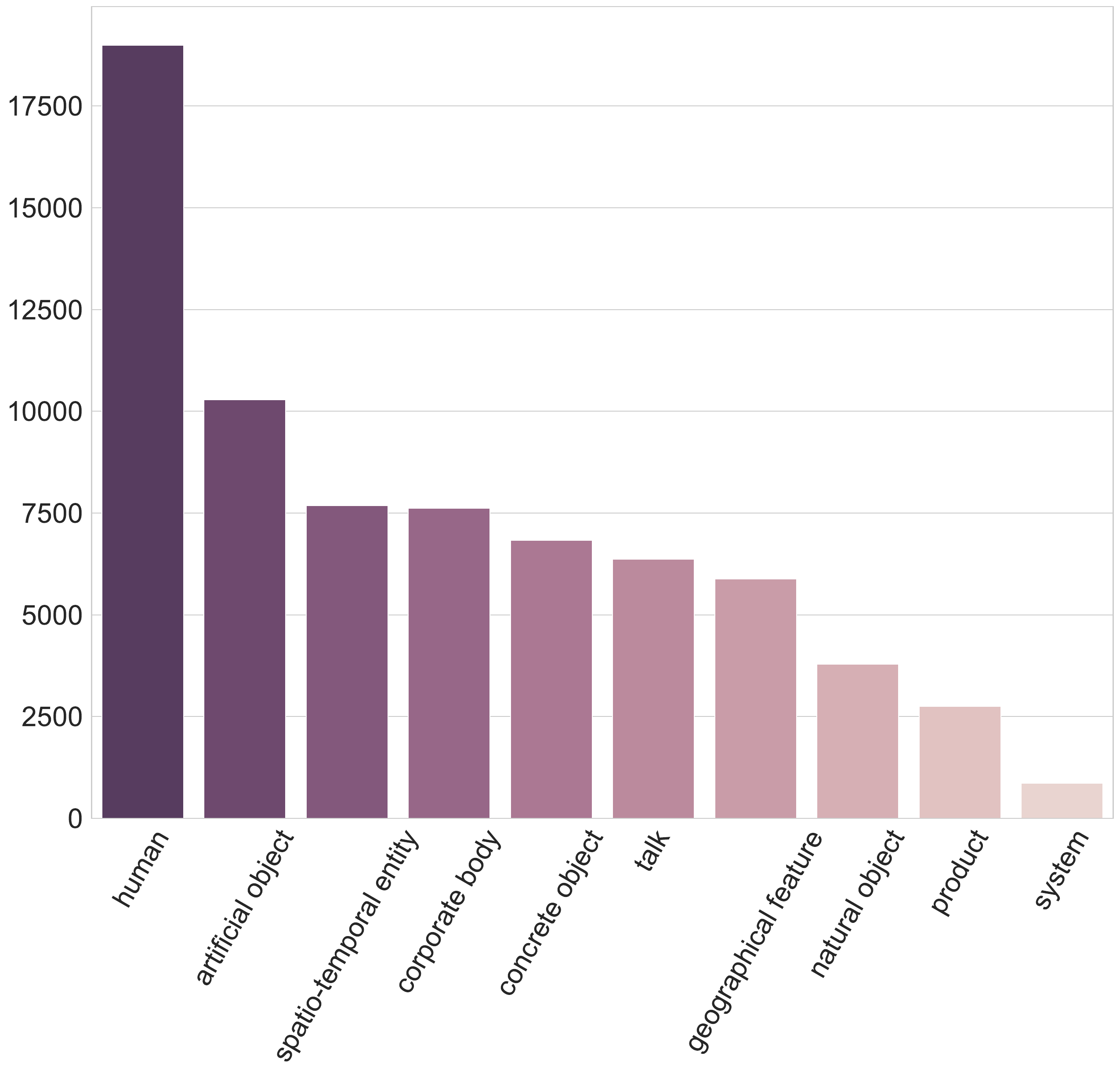}
    \captionsetup{width=0.98\linewidth}
    \caption{Frequency histogram of entity types in Wikidata-based Dataset.}
    \label{fig:wiki_types}
\end{minipage}
\begin{minipage}[t]{.49\textwidth}
  \centering
    \includegraphics[width=0.9\columnwidth]{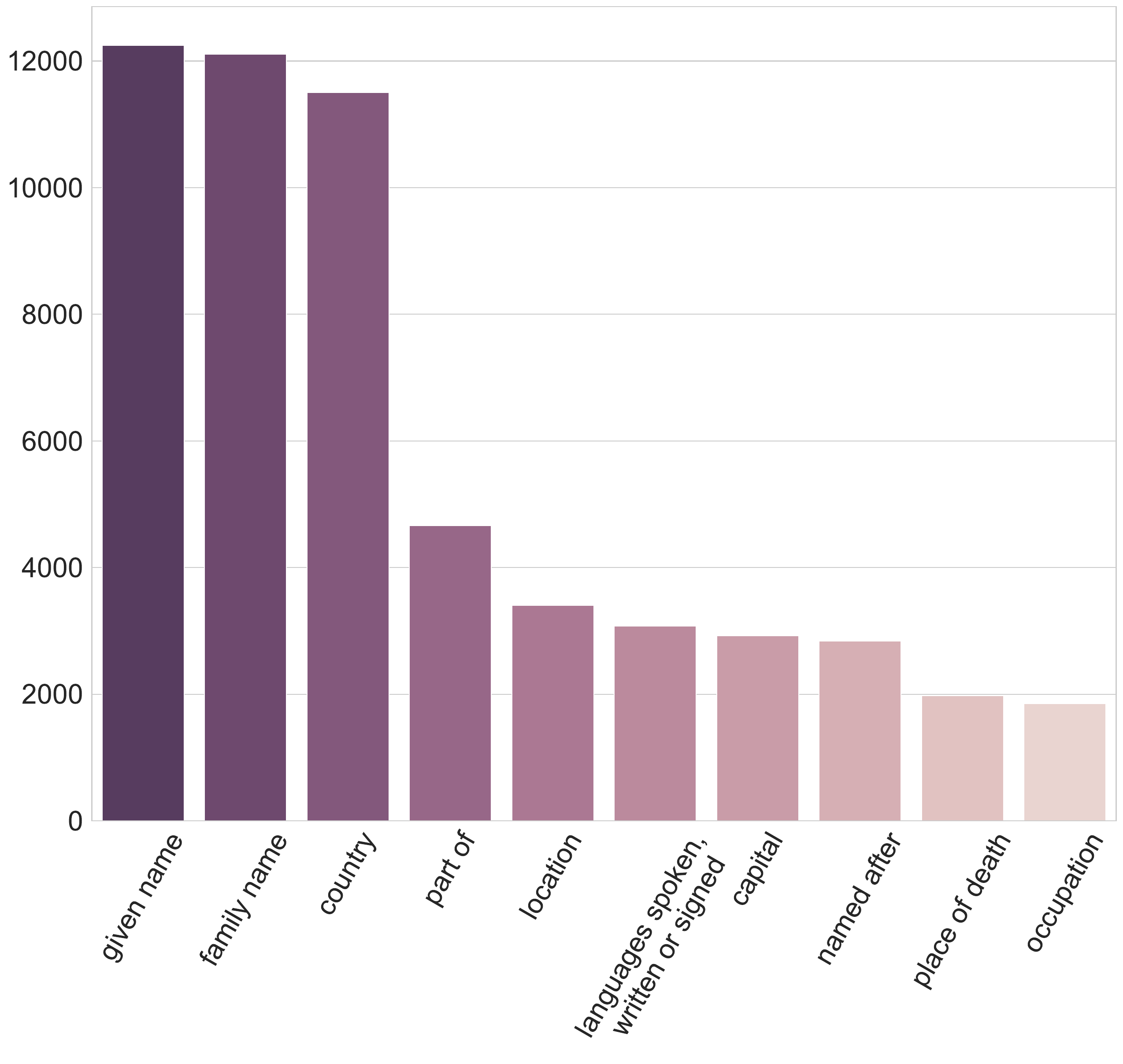}
    \captionsetup{width=0.98\linewidth}
    \caption{Frequency histogram of property keys in Wikidata-based Dataset.}
    \label{fig:wiki_properties}
\end{minipage}
\end{figure}

\section{Ablation Study}\label{appendix:ablation}
%
%
The ablation study conducted on the MuSEE model, with the Wikidata-based dataset, serves as an evaluation of the model's core components: the introduction of special tokens and the Multi-stage parallel generation. 
By comparing the performance of the full MuSEE model against its ablated version, where only the special tokens feature is retained, we aim to dissect the individual contributions of these design choices to the model's overall efficacy.
The ablated version simplifies the output format by eliminating punctuation such as commas, double quotes, and curly brackets, and by converting all entity types and property keys into special tokens. 
This is similar to the reducing output tokens discussed in Sec.~\ref{sec:model}. 
Results from the ablation study, as shown in Table~\ref{tab:ablation_wiki}, reveal significant performance disparities between the complete MuSEE model and its ablated counterpart, particularly when examining metrics across different model sizes (T5-B and T5-L) and evaluation metrics. 
The full MuSEE model markedly outperforms the ablated version across all metrics with notable improvements, underscoring the Multi-stage parallel generation's critical role in enhancing the model's ability to accurately and comprehensively extract entity-related information.
These findings highlight the synergistic effect of the MuSEE model's design elements, demonstrating that both the Reducing output tokens and the Multi-stage parallel generation are pivotal for achieving optimal performance in structured entity extraction tasks. 

\begin{table*}[t]
    \centering
    \caption{Ablation study on Wikidata-based dataset. Each metric is shown in percentage (\%).}
    \label{tab:ablation_wiki}
    \resizebox{0.9\textwidth}{!}
    {
    \begin{tabular}{l|ccc|ccc|ccc}
    \toprule 
        \multirow{2}{*}{Model} & \multicolumn{3}{c}{AESOP-ExactName} & \multicolumn{3}{c}{AESOP-ApproxName} & \multicolumn{3}{c}{AESOP-MultiProp}\\
        & Max & Precision & Recall & Max & Precision & Recall & Max & Precision & Recall\\
        \midrule
         w/o Multi-stage \small{(T5-B)} & 25.19 & 40.87 & 27.64 & 25.75 & 42.14 & 28.26 & 26.93 & 44.49 & 29.72 \\
         \textbf{MuSEE \small{(T5-B)}} & \textbf{44.95} & \textbf{50.63} & \textbf{58.99} & \textbf{45.75} & \textbf{51.57} & \textbf{60.10} & \textbf{46.95} & \textbf{53.00} & \textbf{61.75}\\
         \midrule\midrule
         w/o Multi-stage \small{(T5-L)} & 27.74 & 53.04 & 28.81 & 28.14 & 54.10 & 29.22 & 29.14 & 56.90 & 30.29 \\
         \textbf{MuSEE \small{(T5-L)}} & \textbf{49.35} & \textbf{57.97} & \textbf{59.63} & \textbf{49.89} & \textbf{58.69} & \textbf{60.35} & \textbf{50.94} & \textbf{60.11} & \textbf{61.68}\\
         \bottomrule
    \end{tabular}
    }
\end{table*}

\section{Human Evaluation Criteria and Case Study}\label{appendix:human_eval}
The details for the three human evaluation criteria are shown below:
\begin{itemize}
    \item \textit{Completeness}: Which set of entities includes all relevant entities and has the fewest missing important entities? Which set of entities is more useful for further analysis or processing? Focus on the set that contains less unimportant and/or irrelevant entities.
    \vspace{-2mm}
    \item \textit{Correctness}: Which set of entities more correctly represents the information in the passage? Focus on consistency with the context of the passage. Do extracted properties correctly represent each entity or are there more specific property values available? Are property values useful?
    \vspace{-2mm}
    \item \textit{Hallucinations}: Which set of entities contains less hallucinations? That is, are there any entities or property values that do not exist or cannot be inferred from the text?
\end{itemize}

We provide a case study for the human evaluation analysis comparing the outputs of GenIE (T5-L) and MuSEE (T5-L) given a specific text description. 
MuSEE accurately identifies seven entities, surpassing GenIE's two, thus demonstrating greater completeness.
Additionally, we identify an error in GenIE's output where it incorrectly assigns \textit{Bartolomeo Rastrelli}'s place of death as \textit{Moscow}, in contrast to the actual location, \textit{Saint Petersburg}, which is not referenced in the text. 
This error by GenIE could stem from hallucination, an issue not present in MuSEE's output. 
In this example, it is evident that MuSEE outperforms GenIE in terms of \textit{completeness}, \textit{correctness}, and resistance to \textit{hallucinations}.

\begin{mdframed}[backgroundcolor=gray!20, linewidth=0pt]
\textbf{Text Description:} The ceremonial attire of Elizabeth, Catherine Palace, Tsarskoye Selo; fot. Ivonna Nowicka Elizabeth or Elizaveta Petrovna (; ) reigned as Empress of Russia from 1741 until her death in 1762. She remains one of the most popular Russian monarchs because of her decision not to execute a single person during her reign, her numerous construction projects, and her strong opposition to Prussian policies. The second-eldest daughter of Tsar Peter the Great (), Elizabeth lived through the confused successions of her father's descendants following her half-brother Alexei's death in 1718. The throne first passed to her mother Catherine I of Russia (), then to her nephew Peter II, who died in 1730 and was succeeded by Elizabeth's first cousin Anna (). After the brief rule of Anna's infant great-nephew, Ivan VI, Elizabeth seized the throne with the military's support and declared her own nephew, the future Peter III, her heir. During her reign Elizabeth continued the policies of her father and brought about a remarkable Age of Enlightenment in Russia. Her domestic policies allowed the nobles to gain dominance in local government while shortening their terms of service to the state. She encouraged Mikhail Lomonosov's foundation of the University of Moscow, the highest-ranking Russian educational institution. Her court became one of the most splendid in all Europe, especially regarding architecture: she modernised Russia's roads, encouraged Ivan Shuvalov's foundation of the Imperial Academy of Arts, and financed grandiose Baroque projects of her favourite architect, Bartolomeo Rastrelli, particularly in Peterhof Palace.
\end{mdframed}

\textbf{GenIE (T5-L):}
\begin{lstlisting}[language=json]
{
    "0": {
        "entity name": "Bartolomeo Rastrelli",
        "place of death": "Moscow",
        "given name": "Bartolomeo",
        "family name": "Rastrelli",
        "languages spoken, written or signed": "Russian",
        "country": "Russia",
        "type": "human"
    },
    "1": {
        "entity name": "Catherine I of Russia",
        "given name": "Catherine",
        "languages spoken, written or signed": "Russian",
        "type": "human"
    }
}
\end{lstlisting}
\textbf{MuSEE (T5-L):}
\begin{lstlisting}[language=json]
{
    "0": {
        "type": "human",
        "entity name": "Peter the Great",
        "given name": "Peter"
    },
    "1": {
        "type": "human",
        "entity name": "Bartolomeo Rastrelli",
        "family name": "Rastrelli",
        "given name": "Bartolomeo"
    },
    "2": {
        "type": "human",
        "entity name": "Catherine I of Russia",
        "given name": "Catherine"
    },
    "3": {
        "type": "human",
        "entity name": "Peter III",
        "given name": "Peter"
    },
    "4": {
        "type": "human",
        "entity name": "Ivan Shuvalov",
        "family name": "Shuvalov",
        "given name": "Ivan",
        "languages spoken, written or signed": "Russian"
    },
    "5": {
        "type": "concrete object",
        "entity name": "Peterhof Palace",
        "country": "Russia"
    },
    "6": {
        "type": "human",
        "entity name": "Mikhail Lomonosov",
        "family name": "Lomonosov",
        "given name": "Mikhail",
        "languages spoken, written or signed": "Russian"
    }
}
\end{lstlisting}

\begin{figure*}[ht]
    \centering
    \includegraphics[width=0.9\linewidth]{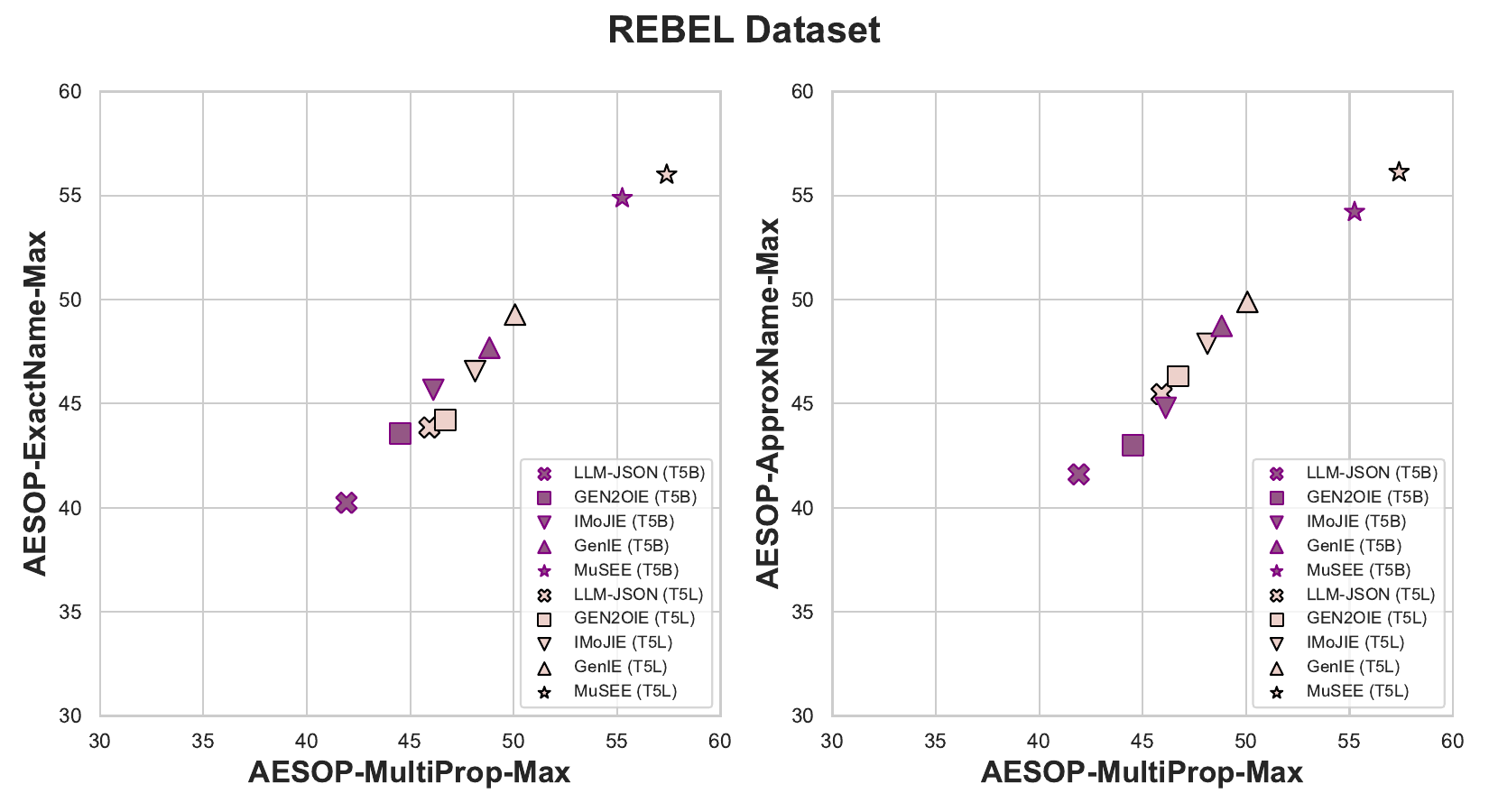}
    \caption{Metric correlation analysis on the REBEL dataset.}
    \label{fig:metric_REBEL}
\end{figure*}

\begin{figure*}[ht]
    \centering
    \includegraphics[width=0.9\linewidth]{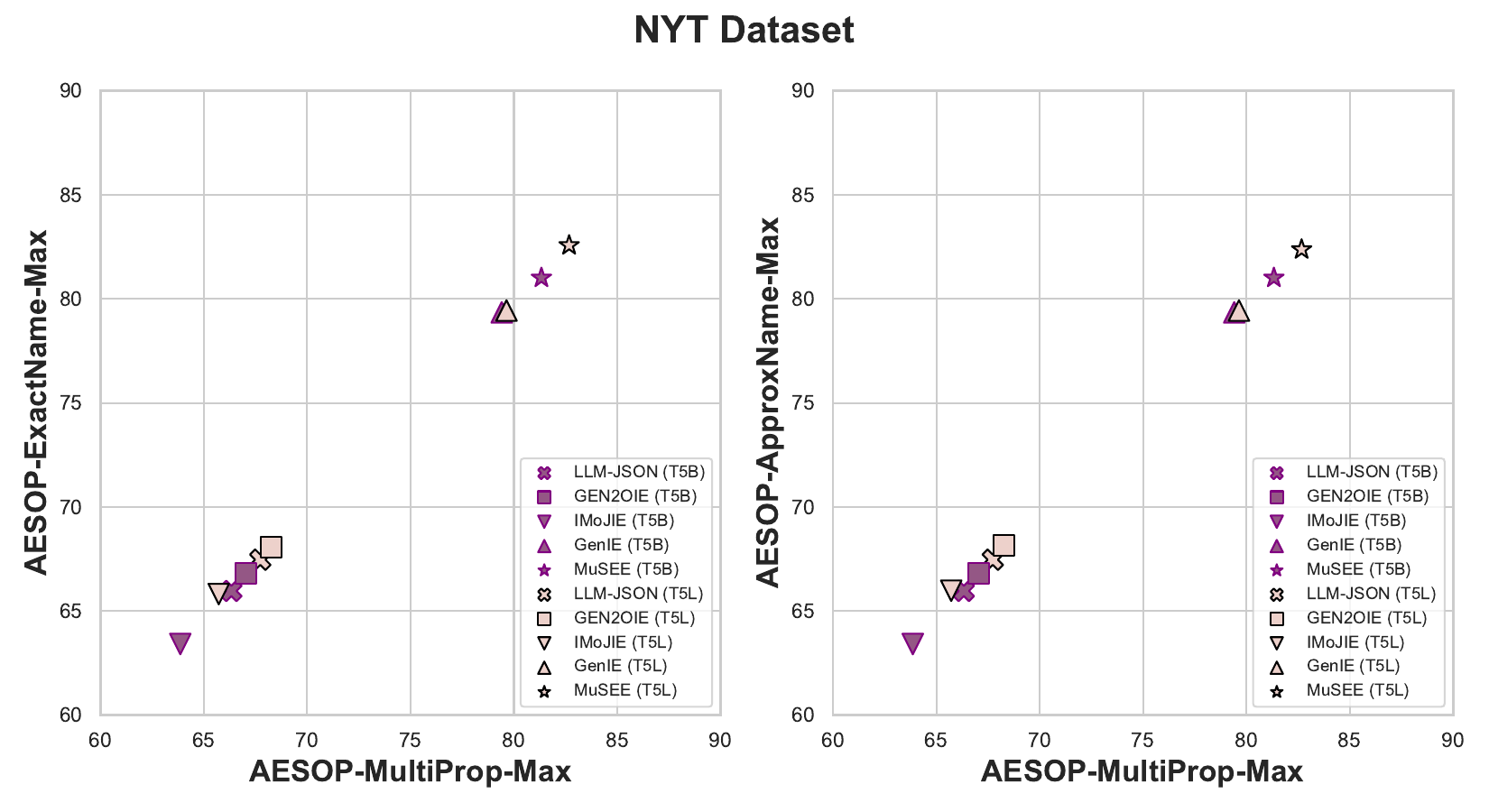}
    \caption{Metric correlation analysis on the NYT dataset.}
    \label{fig:metric_NYT}
\end{figure*}

\begin{figure*}[ht]
    \centering
    \includegraphics[width=0.9\linewidth]{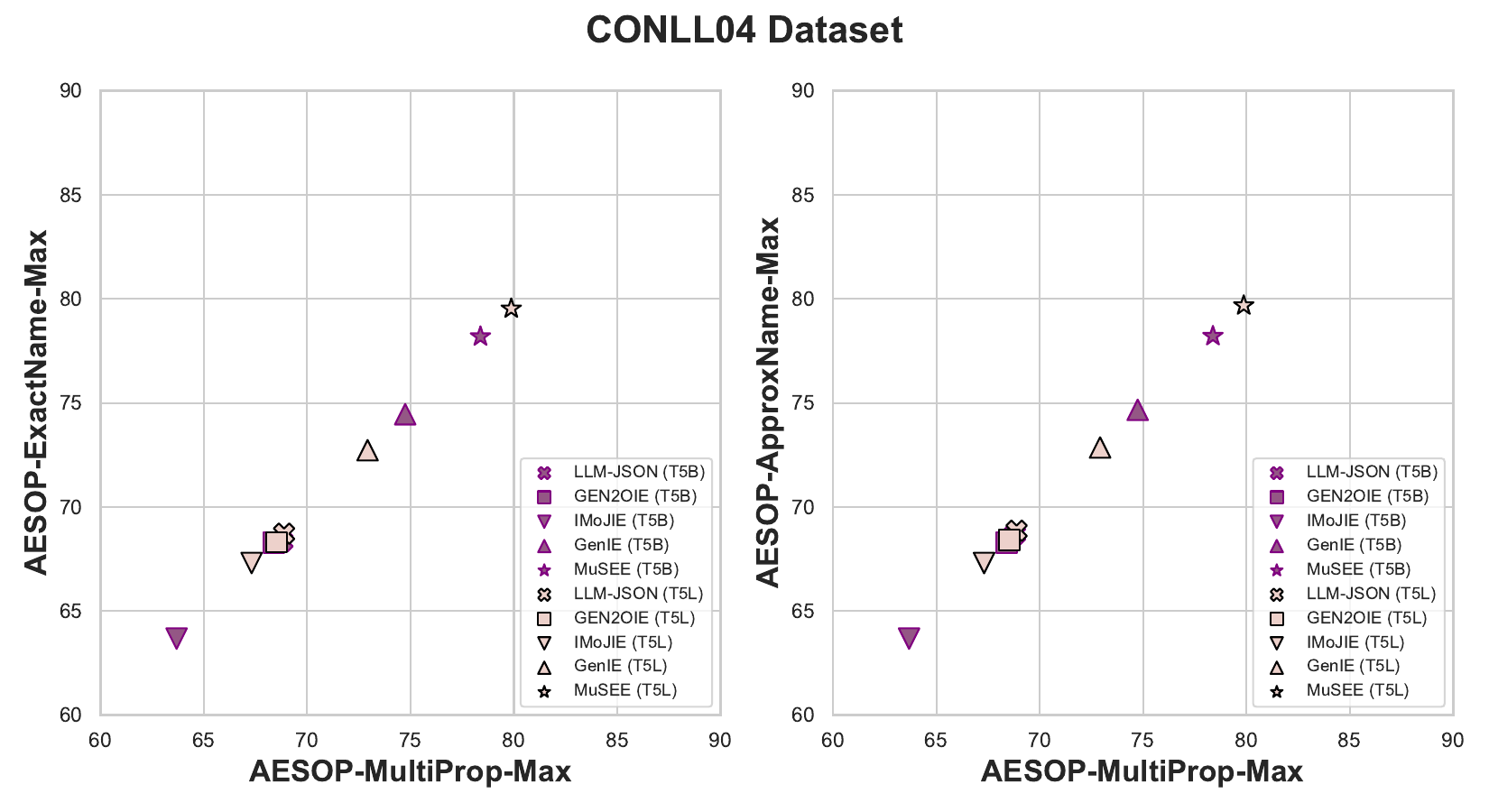}
    \caption{Metric correlation analysis on the CONLL04.}
    \label{fig:metric_CONLL04}
\end{figure*}

\begin{figure*}[ht]
    \centering
    \includegraphics[width=0.9\linewidth]{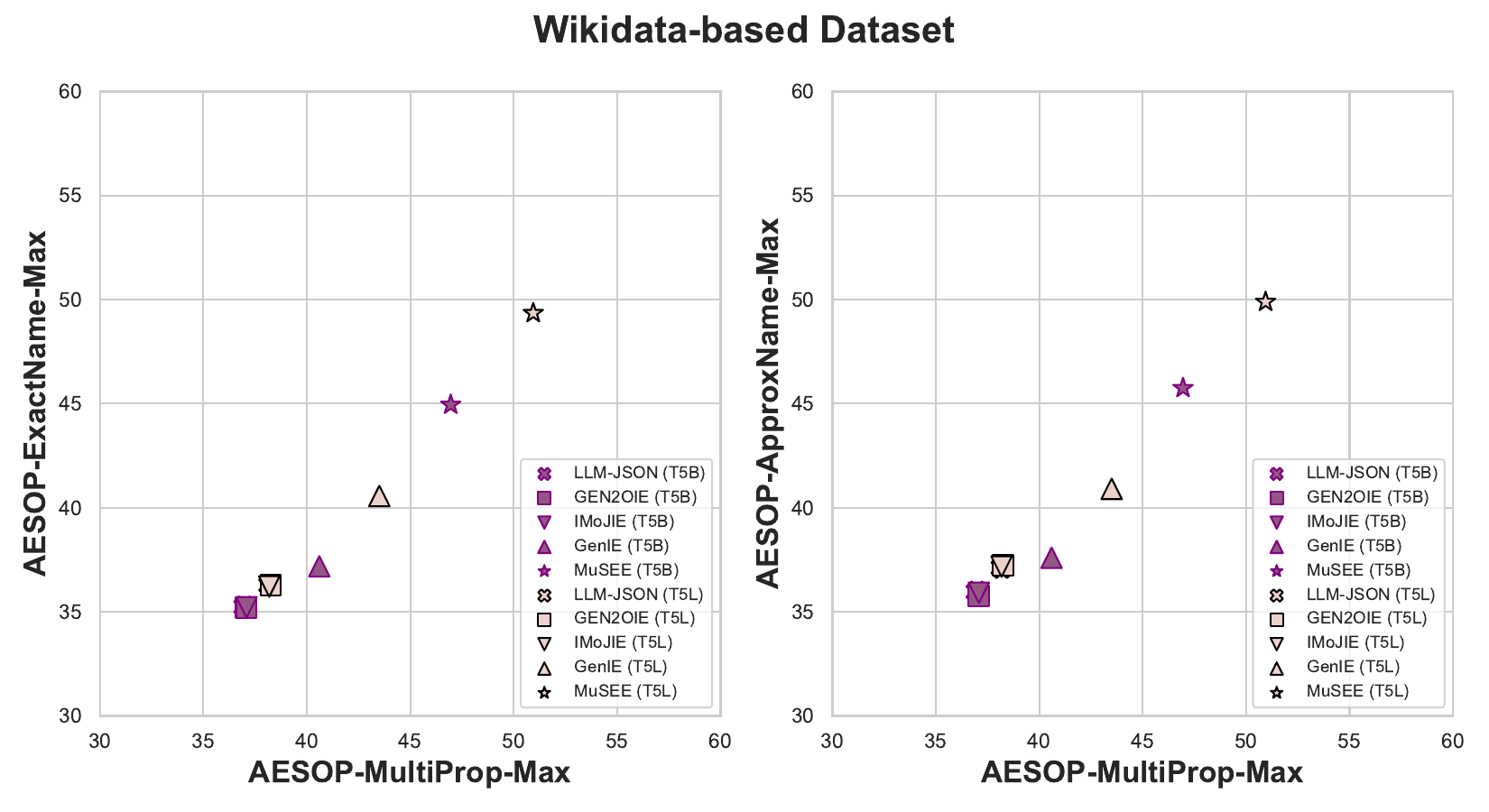}
    \caption{Metric correlation analysis on the Wikidata-based dataset.}
    \label{fig:metric_wiki}
\end{figure*}


\section{Metric Correlation Analysis}
We show the correlation analysis between AESOP metric variants across all models on all four datasets, shown in Fig.~\ref{fig:metric_REBEL}, Fig.~\ref{fig:metric_NYT}, Fig.~\ref{fig:metric_CONLL04}, and Fig.~\ref{fig:metric_wiki}, respectively.
Specifically, we focus on the correlation analysis of different variants based on entity assignment variants in Phase 1 of AESOP, as described in Sec.~\ref{sec:prelim}.
For Phase 2, the ``Max'' normalization method is employed by default.
Observations for the other two normalization variants are similar.
In the associated figures, AESOP-MultiProp-Max is uniformly used as the x-axis measure, while AESOP-ExactName-Max or AESOP-ApproxName-Max serve as the y-axis metrics. 
The scatter plots in all figures tend to cluster near the diagonal, indicating a robust correlation among the various metric variants we have introduced.


\end{document}